\documentclass[10pt,twocolumn,letterpaper]{article}

\usepackage[pagenumbers]{cvpr} % To force page numbers, e.g. for an arXiv version

\definecolor{cvprblue}{rgb}{0.21,0.49,0.74}
\definecolor{mygray}{RGB}{242,242,242}

\definecolor{fig3_blue}{rgb}{.4,.8039,1}

\definecolor{myblue}{rgb}{.874,.956,.992}

\definecolor{table_blue}{RGB}{47,127,193}
\definecolor{table_gray}{RGB}{128,138,135}

\definecolor{visred}{RGB}{255,217,217}
\definecolor{visgreen}{RGB}{229,241,221}
\definecolor{visviolet}{RGB}{240,234,245}

\newcommand{\newgain}[1]{\textcolor{table_gray}{(+{#1})}}
\newcommand{\newdrop}[1]{\textcolor{table_blue}{(-{#1})}}

\definecolor{table_purple}{RGB}{142,139,254}
\definecolor{table_gray_2}{RGB}{169,184,198}
\newcommand{\gain}[1]{\textcolor{table_purple}{(+{#1})}}
\newcommand{\drop}[1]{\textcolor{table_gray_2}{(-{#1})}}

\usepackage[pagebackref,breaklinks,colorlinks,allcolors=cvprblue]{hyperref}

\usepackage{url}
\usepackage{booktabs} % for table
\usepackage{multirow} % for multirow table
\usepackage{makecell} % for // in table
\usepackage{xcolor}   % for color
\usepackage{amssymb}  % for \checkmark and so on
\usepackage{colortbl}   % for \cellcolor 
\usepackage{pifont} % for ding
\usepackage{threeparttable} % For table notes

\usepackage{cuted} 
\usepackage{capt-of}

\def\paperID{5967} % *** Enter the Paper ID here
\def\confName{CVPR}
\def\confYear{2026}

\title{AMap: Distilling Future Priors for Ahead-Aware Online HD Map Construction}

\author{
    Ruikai Li\textsuperscript{\rm 1,2}$^{\ast}$,
    Xinrun Li\textsuperscript{\rm 3}$^{\ast}$,
    Mengwei Xie\textsuperscript{\rm 2},
    Hao Shan\textsuperscript{\rm 1}, \\
    Shoumeng Qiu\textsuperscript{\rm 1},
    Xinyuan Chang\textsuperscript{\rm 2},
    Yizhe Fan\textsuperscript{\rm 1},
    Feng Xiong\textsuperscript{\rm 2},
    Han Jiang\textsuperscript{\rm 1},\\
    Yilong Ren\textsuperscript{\rm 1},
    Haiyang Yu\textsuperscript{\rm 1},
    Mu Xu\textsuperscript{\rm 2},
    Yang Long\textsuperscript{\rm 4},
    Varun Ojha\textsuperscript{\rm 3},
    Zhiyong Cui\textsuperscript{\rm 1}$^{\dagger}$\\
    \textsuperscript{\rm 1} State Key Lab of Intelligent Transportation System, China 
    \textsuperscript{\rm 2} Amap, Alibaba Group, China \\
    \textsuperscript{\rm 3} Newcastle University, England
    \textsuperscript{\rm 4} Durham University, England \\
    {\small $^{\ast}$Equal contribution \qquad $^{\dagger}$Corresponding author}
}
\begin{document}
%\maketitle
\twocolumn[{%
    \renewcommand\twocolumn[1][]{#1}%
    \maketitle
    \vspace{-0.5in}
    \begin{center}
        \centering
        \includegraphics[width=1.0\linewidth]{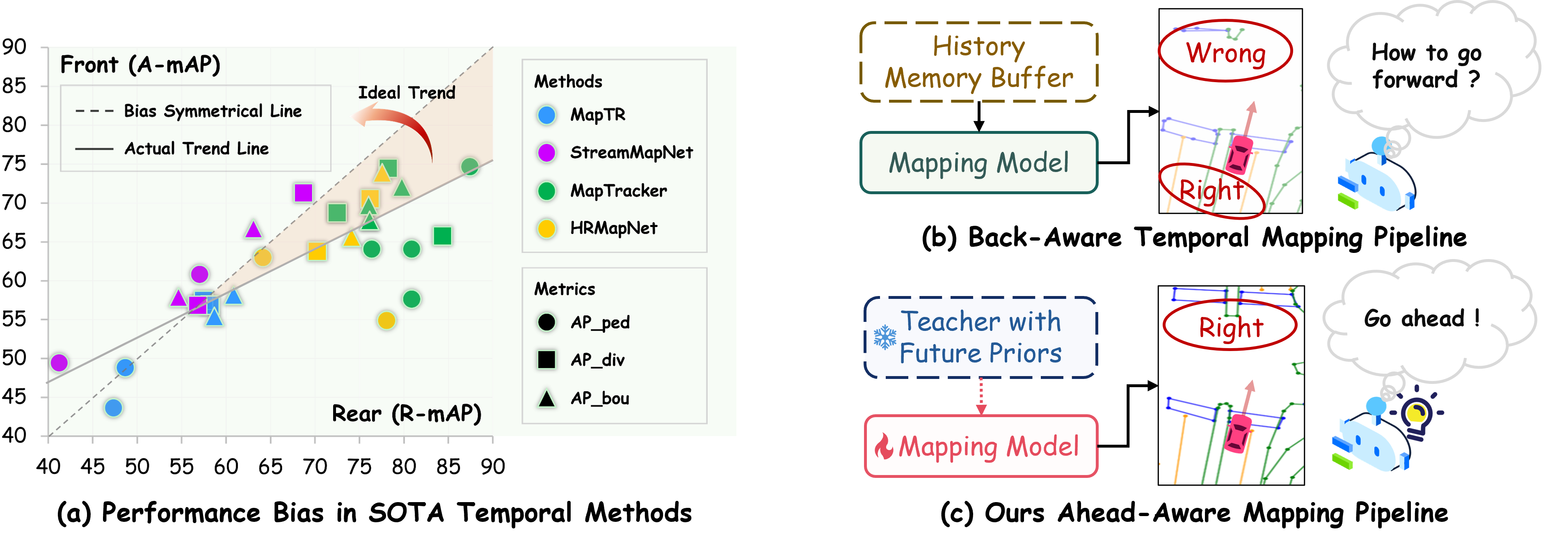}
        \captionof{figure}{\textbf{(a) Performance Bias in SOTA Temporal Methods.} Current state-of-the-art temporal mapping methods exhibit a severely weeker mAP in front region (A-mAP) than it in rear views (R-mAP), as illustrated in the orange-shaded region. \textbf{(b)~\&~(c) Back-Aware Pipeline \textit{vs}. Our Ahead-aware Pipeline.} While the (b)~Back-Aware mapping paradigm enhances perception in the vehicle's rearward area, this comes at the cost of limited gains in forward spatial awareness. In contrast, our paradigm~(c) effectively achieves forward perception gains by acquiring future priors through knowledge distillation.}
        \label{fig:teaser}
    \end{center}
    %\vspace{0.3cm}
    \vspace{-0.05in}
}]

\vspace{-3mm}
\begin{abstract}
Online High-Definition (HD) map construction is pivotal for autonomous driving. While recent approaches leverage historical temporal fusion to improve performance, we identify a critical safety flaw in this paradigm: it is inherently ``spatially backward-looking." These methods predominantly enhance map reconstruction in traversed areas, offering minimal improvement for the unseen road ahead. Crucially, our analysis of downstream planning tasks reveals a severe asymmetry: while rearward perception errors are often tolerable, inaccuracies in the forward region directly precipitate hazardous driving maneuvers. To bridge this safety gap, we propose AMap, a novel framework for Ahead-aware online HD Mapping. We pioneer a ``distill-from-future" paradigm, where a teacher model with privileged access to future temporal contexts guides a lightweight student model restricted to the current frame. This process implicitly compresses prospective knowledge into the student model, endowing it with ``look-ahead" capabilities at zero inference-time cost. Technically, we introduce a Multi-Level BEV Distillation strategy with spatial masking and an Asymmetric Query Adaptation module to effectively transfer future-aware representations to the student's static queries. Extensive experiments on the nuScenes and Argoverse 2 benchmark demonstrate that AMap significantly enhances current-frame perception. Most notably, it outperforms state-of-the-art temporal models in critical forward regions while maintaining the efficiency of single current frame inference.

% \TODO{To be Regenerated after intro polished}
% High Definition map construction is a critical task for autonomous driving.While recent methods leverage temporal fusion of historical frames to improve accuracy, we identify a fundamental limitation: this backward looking paradigm enhances map reconstruction primarily behind the ego vehicle, offering minimal improvement for the critical forward region that is most relevant to driving decisions. To address this issue, we propose AMap, a novel framework for Ahead Aware online HD Map construction via future priors distillation. Our core idea is to transfer knowledge from a powerful teacher model that incorporates future temporal contexts into a lightweight student model restricted to the current frame. This process implicitly compresses future aware representations into the student, enabling it to perceive the road ahead more accurately without any computational overhead during inference. Specifically, we introduce two key technical components: a Multi Level Temporal BEV Distillation module that transfers future context across different feature resolutions, and an Asymmetric Query Distillation module that adaptively aligns the teacher's dynamic instance queries with the student's static ones. Extensive experiments on the nuScenes dataset demonstrate that our approach significantly enhances the perceptual accuracy of a single frame model, particularly in forward regions, and even surpasses several state of the art temporal models in frontal perception while maintaining real time efficiency. \TODO{review}
\end{abstract}   

\vspace{-0.3in}
\section{Introduction}
\label{sec:intro}

% \TODO{Xinrun: 
% 1. We are the first to identify and systematically analyze the "backward-looking" flaw in current temporal-fusion mappers. To quantify this, we introduce a new benchmark and a novel forward-aware metric designed to evaluate perception accuracy in the critical, unseen areas ahead of the vehicle.
% 2. We propose AMap, a novel ahead-aware knowledge distillation framework that endows a single-frame model with "look-ahead" capabilities at zero inference-time cost. Our framework introduces several components for this task, including BEV masking, multi-level BEV feature distillation, and query matching transfer to effectively distill prospective knowledge.
% 3. Comprehensive evaluations show our method achieves state-of-the-art performance on the proposed forward-aware metrics. Furthermore, we validate AMap's effectiveness in downstream planning tasks, demonstrating a significant reduction in critical errors caused by forward-region misperception.
% }

% \TODO{1. all model ahead-aware bad!!! 2. a fig to highlight all model in ahead-aware }

HD Map is indispensable for autonomous driving systems, providing precise and stable semantic representations of static environments. However, traditional offline HD map is plagued by high costs and poor scalability. To overcome these limitations, recent research has spurred the development of online HD mapping techniques \cite{li2022hdmapnet,liu2023vectormapnet,qiao2023bemapnet,ding2023pivotnet,zhang2023mapvr,liao2025maptrv2,liu2024mgmap,zhang2024gemap,liu2024mapqr}. These methods seek to infer vectorized representations of map elements directly and in real-time from the vehicle's onboard sensory data, and have demonstrated promising performance in complex autonomous driving scenarios. 
A prevailing trend among state-of-the-art online mapping methods is the integration of historical temporal information \cite{liao2022maptr,yuan2024streammapnet,li2024dtclmapper,chen2024maptracker,yang2025histrackmap}. By processing sequential data, these approaches leverage historical context to enhance the perception of the current frame. This effectively mitigates current-frame limitations, such as occlusion, to yield more consistent and complete map outputs.

Nevertheless, this reliance on historical data introduces a critical and largely unaddressed flaw: ``backward-looking" bias, as illustrated in Fig.~\ref{fig:teaser}~(a) and (b). Through a detailed analysis of existing state-of-the-art temporal models on the nuScenes \cite{caesar2020nuscenes} benchmark, we find that their performance gains are not uniform. To quantify this, we introduce a novel ahead-aware evaluation metric (see Sec.~\ref{subsec:3-1-A-mAP}). Our analysis reveals a stark asymmetry: the temporal fusion primarily enhances perception accuracy in areas where the ego-vehicle has already traversed, while its capability in the critical, unseen road ahead remains largely unimproved.

% Nevertheless, we identify a critical flaw in this historical-fusion paradigm: it is intrinsically a backward-looking process. The temporal context it leverages is overwhelmingly concentrated on areas the ego-vehicle has already traversed, offering little for the critical unseen road ahead. This inherent bias creates a stark asymmetry: the model's capability is enhanced predominantly for the areas behind the vehicle, while its perception of the forward region, which is paramount for driving decisions, remains largely unimproved.

% This key defect creates a fundamental mismatch with an autonomous driving system's operational priorities. Since planning and decision-making rely more critically on accurately understanding the upcoming road geometry than on the passed areas, a rearview-heavy perception model risks failure in downstream tasks, posing a severe safety hazard. As depicted in Figure \TODO{vis}, although the mapping model performs well behind the vehicle, its predictive errors for the road ahead can cause the autonomous system to execute dangerous maneuvers.

This performance bias creates a fundamental mismatch with an autonomous driving system's operational priorities. Planning and decision-making systems rely far more critically on an accurate understanding of the upcoming road geometry than on areas already passed. Our preliminary experiments further illuminate this: we observe that masking out the rearward perception output often leads to a less severe degradation in downstream planning performance, and in some cases, even a slight improvement, compared to masking out the forward perception. This stark contrast underscores the differential importance of accurate perception in these distinct regions. Such findings lead us to a critical research question: \textit{Is it feasible to design a lightweight, plug-and-play module to eliminate the forward-perception deficit without requiring additional architectural changes or inference cost?}
% {can we eliminate this forward-perception deficit without incurring additional inference-time costs or requiring complex architectural changes?}
%As depicted in Figure \TODO{vis}, while the map behind the vehicle is accurate, predictive errors in the forward region can cause the planning module to execute unsafe or erroneous maneuvers, posing a severe safety hazard.

To bridge this gap, we propose AMap, a novel Ahead-aware Mapping framework based on knowledge distillation. Our core insight is to leverage future priors, which are readily available during the training phase, as a powerful supervisory signal to guide a standard current-frame (online) model. As illustrated in Fig.~\ref{fig:teaser}~(c), we employ a ``distill from the future" strategy. We first train a powerful teacher model that incorporates future temporal context. Then, we distill its informative, future-aware representations into a lightweight student model that is restricted to only the current frame at inference. This process implicitly compresses prospective knowledge into the student, endowing it with a ``look-ahead" capability without processing any future (or even historical) frames during deployment.

% To bridge this gap, we propose AMap, a novel Ahead-Aware knowledge distillation framework. Our core insight is to leverage future priors, which are readily available during training, as a rich supervisory signal for guiding a purely online model. As illustrated in Fig.~\ref{fig:teaser}, we distill informative future-aware representations from a powerful teacher model that incorporates future temporal contexts into a lightweight student model restricted to the current frame. This process implicitly compresses prospective knowledge into the student, empowering it with a “look-ahead” capability during inference and achieving significant performance gains precisely where they matter most: in the vehicle’s forward path.

This distillation task, however, presents unique challenges. Standard distillation methods are not directly applicable, particularly for the dynamic, set-based predictions of vectorized mapping. The student (online) and teacher (future-aware) models operate on different contexts, creating an ``asymmetric" setting for distillation, especially at the query-decoder level. To address this, we introduce three technical components: Bird's Eye View~(BEV) masking, multi-level BEV feature distillation, and query matching transfer to effectively transfer knowledge from the teacher's decoder to the student, despite the contextual asymmetry.
% Comprehensive evaluations on the nuScenes \cite{caesar2020nuscenes} and Argoverse 2\cite{wilson2023argoverse} benchmark demonstrate that AMap markedly improves the perceptual accuracy of the current-frame model, particularly in the critical forward regions. Impressively, our current-frame model trained with AMap achieves state-of-the-art performance on both the overall mAP and our proposed forward-aware metrics, even outperforming several advanced temporal models (with R18 backbone) while maintaining the high efficiency of current-frame inference.
Our principal contributions are threefold:
\begin{itemize}

\item 
We are the first to identify and systematically analyze the ``backward-looking" flaw in current temporal-fusion mappers. To quantify this, we introduce a new benchmark and a novel forward-aware metric designed to evaluate perception accuracy in the critical, unseen areas ahead of the vehicle.
% We propose \textbf{AMap}, a novel ahead-aware online HD mapping framework that leverages future priors via knowledge distillation to endow a single-frame model with look-ahead capability, without incurring inference-time cost.

\item 
We propose AMap, a novel ahead-aware knowledge distillation framework that pioneers a ``distill-from-future" paradigm. By leveraging a teacher model with privileged access to future temporal contexts, we explicitly transfer prospective knowledge into a current-frame student. This endows the model with ``look-ahead" capabilities at zero inference-time cost, supported by our proposed BEV masking, multi-level BEV feature distillation, and query matching transfer modules.
% We introduce two key technical components within AMap: multi-level distillation strategy for BEV features that transfers future-aware knowledge from the teacher to the student across different feature resolutions and granularities, and an \TODO{XXX} asymmetric query adaptation module to effectively transfer knowledge to the student's decoder.

\item 
Comprehensive evaluations on the nuScenes \cite{caesar2020nuscenes} and Argoverse 2 \cite{wilson2023argoverse} benchmark demonstrate that our approach markedly improves the perceptual accuracy of the current-frame model, particularly in critical forward areas. Impressively, it outperforms several advanced temporal models in frontal perception accuracy while maintaining the efficiency of single current frame inference.

\end{itemize}

\vspace{-0.1in}
\section{Related Work}
\label{sec:related}

\noindent\textbf{Online HD Map Construction.} While indispensable for autonomous driving, traditional HD maps suffer from high costs and an inability to handle dynamic scenes, motivating the development of online mapping techniques. Early online methods \cite{philion2020lss,liu2022bevfusion,li2022hdmapnet,jiang2024pmapnet} centered on generating BEV semantic maps. However, this resulted in suboptimal pixel output for downstream planning. This limitation was overcome by subsequent end-to-end vectorized paradigms \cite{liu2023vectormapnet,liao2022maptr,liao2025maptrv2,liu2024mapqr,zhang2024hrmapnet,hao2024mapbench,ding2023pivotnet,wan2025sp2t,liu2024mgmap,zhang2024gemap,zhang2023mapvr,chang2025mapdr,yuan2025unimapgen,yan2025mapkd,xie2025seqgrowgraph,wu2025interactionmap,liu2025UIGenMap,dong2025damap,qiu2025MapGR,shan2025stability,wan2025nips}, which produce structured, instance-level map elements for immediate use. The frontier of research now lies in exploiting temporal information, where the integration of historical frames has led to state-of-the-art accuracy by enforcing temporal consistency \cite{liao2022maptr,yuan2024streammapnet,li2024dtclmapper,chen2024maptracker,yang2025histrackmap}. However, these temporal fusion methods rely predominantly on historical information from areas in which the vehicle has already passed. Consequently, while they achieve impressive overall accuracy, the resulting performance gain offers limited improvement in perceiving the forward region, which is arguably more critical for planning and safety. Furthermore, the introduction of temporal fusion modules often brings substantial parameter increases and significant computational overhead, hindering real-time application. This paper addresses these limitations by introducing a novel distillation framework that leverages future priors to explicitly enhance ahead-aware perception. Our approach not only shifts the focus from a backward-looking to a forward-aware paradigm, but also achieves this without introducing any additional inference-time cost.

\noindent\textbf{Knowledge Distillation.} Knowledge Distillation (KD) compresses knowledge from a powerful teacher model into an efficient student model. This strategy is prevalent in autonomous driving perception, including 3D object detection \cite{chen2022bevdistill,zhou2023unidistill,zhao2024simdistill,kwon2025memdistill}, online HD mapping \cite{hao2024mapdistill}, topology understanding \cite{li2025onestage}, and 3D occupancy prediction \cite{ke2025occ1,zhang2024radocc}. Typically, for information that is infeasible during deployment due to cost or latency, KD leverages it as privileged supervision to guide a lightweight student model via an implicit training-time transfer, thereby enhancing its capability without inference overhead. A canonical paradigm in this domain involves distilling cross-modal knowledge from a multi-modal teacher into a uni-modal student branch. Beyond modality transfer, KD also serves as a robust source of supervision in the form of pseudo-labels \cite{lilja2025semi} or intermediate representations to facilitate model training \cite{hao2025safemap}. In this work, we extend the application of KD to the temporal dimension. Specifically, we propose distilling knowledge from a teacher model that is privileged with future temporal information into an efficient student model, thereby instilling anticipatory perception capabilities for online mapping.

\noindent\textbf{Temporal Modeling for Perception.}
Effective temporal modeling is crucial for a coherent and robust autonomous driving perception system. Existing methods can be broadly categorized by the temporal context they leverage. A dominant line of research focuses on historical information, integrating past sensor data to resolve perceptual ambiguities in the present frame. This approach, evident in BEV-based 3D object detection \cite{li2024bevformer,huang2022bevdet4d} and online map construction \cite{yuan2024streammapnet,chen2024maptracker}, primarily aims at enhancing temporal consistency, smoothing outputs, and mitigating occlusions by accumulating information from already-observed areas. In contrast, another body of work engages with future information, most prominently in the field of motion forecasting \cite{li2025occscene,zhang2025iros,wang2024drivedreamer,li2025uniscenev2}. Here, the goal is to anticipate the future states of dynamic agents, relying heavily on understanding their past trajectories and interactions. Our work introduces a novel perspective on temporal modeling by leveraging future information not for forecasting dynamic agents, but for enhancing the perception of the static environment itself. We demonstrate that future frames constitute a powerful source of privileged information for learning a more accurate and anticipatory representation of the road geometry, effectively addressing the spatial misalignment inherent in historical-fusion methods for online HD mapping.

\begin{figure}[t]
\centering
\setlength{\abovecaptionskip}{0.05cm}
\setlength{\belowcaptionskip}{-0.1cm}
\includegraphics[width=0.90\linewidth]{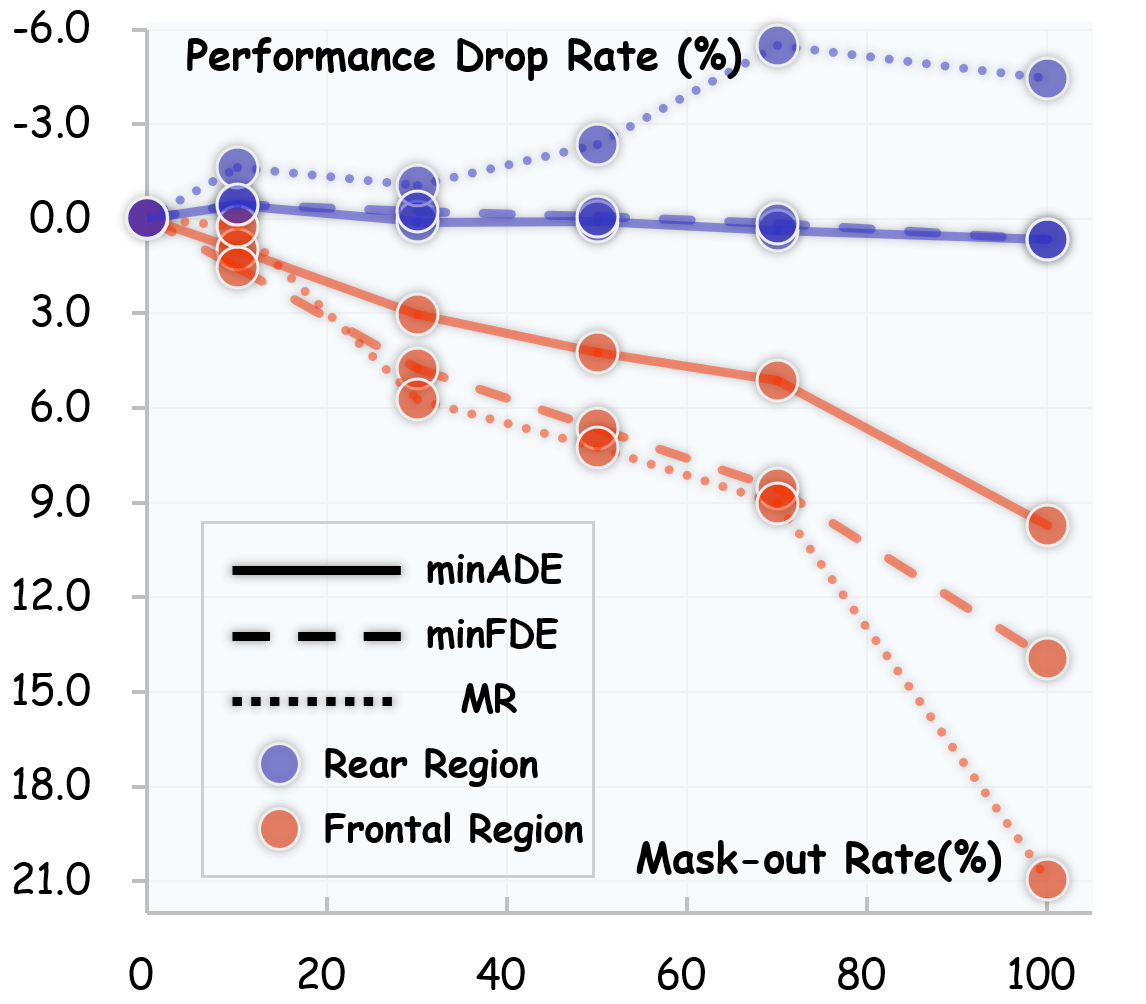}
\caption{\textbf{Impact of directional masking the outputs of the online HD mapper on downstream trajectory prediction.} As the Forward Mask ratio increases (red line), metrics such as minADE, minFDE, and MR deteriorate significantly, whereas the Backward Mask (blue line) has negligible impact. This empirical evidence demonstrates that downstream tasks are critically dependent on the ahead-aware mapping outputs. }
\label{fig:downstream_bias}
\vspace{-0.2in}
\end{figure}

\section{Preliminary Experiments}
\label{sec:pre-exp}

To quantitatively assess how the performance of online mapping models in forward and backward spatial domains influences downstream tasks, we design a downstream task-driven evaluation framework. Our approach is motivated by findings from~\cite{zhang2025iros}, which revealed that trajectory prediction accuracy is highly dependent on the quality of online maps. Building on this, we posit that if the contributions of forward and backward map regions to a downstream task are symmetric, then performance degradation in either region should yield comparable effects. In contrast, asymmetric impacts would indicate a spatial bias in the downstream model’s reliance on map information.

%\vspace{-2mm}
\begin{table}[htbp]
\centering
\caption{\textbf{Downstream Sensitivity to Forward Degradation.}}
\vspace{-2mm}
\fontsize{7pt}{9pt}\selectfont
\label{tab:pre-downstream_forward}
\begin{tabular}[t]{c|ccc}
\toprule
\rowcolor{mygray}
 & \multicolumn{3}{c}{\bf Downstream Task Metric} \\
\cline{2-4}
\rowcolor{mygray}
\multirow{-2}{*}{\makecell[c]{\bf Mask \\ \bf Ratio}} & 
\multirow{-1}{*}{\bf minADE$\downarrow$} &
\multirow{-1}{*}{\bf minFDE$\downarrow$} &
\multirow{-1}{*}{\bf MR$\downarrow$} \\
\midrule
 - & 0.3856 & 0.7919 & 0.0854 \\
\midrule

%5\% & 0.3875 & 0.7998 & 0.0851 \\
10\% & 0.3893 \newgain{0.96\%} & 0.8040 \newgain{1.53\%} & 0.0856 \newgain{0.23\%} \\
%15\% & 0.3906 & 0.8099 & 0.0876 \\
%25\% & 0.3956 & 0.8246 & 0.0898 \\
30\% & 0.3973 \newgain{3.03\%} & 0.8296 \newgain{4.76\%} & 0.0903 \newgain{5.74\%} \\
%40\% & 0.4002 & 0.8395 & 0.0913 \\
50\% & 0.4019 \newgain{4.23\%} & 0.8446 \newgain{6.65\%} & 0.0916 \newgain{7.26\%} \\
%60\% & 0.4035 & 0.8525 & 0.0940 \\
70\% & 0.4054 \newgain{5.13\%} & 0.8595 \newgain{8.54\%} & 0.0931 \newgain{9.02\%} \\
%80\% & 0.4071 & 0.8637 & 0.0949 \\
%90\% & 0.4153 & 0.8835 & 0.1004 \\
100\% & 0.4231 \newgain{9.73\%} & 0.9024 \newgain{13.95\%} & 0.1033 \newgain{20.96\%}  \\
\bottomrule
\end{tabular}
%\vspace{-0.2in}
\end{table}
%\vspace{-2mm}
\begin{table}[htbp]
\centering
\caption{\textbf{Downstream Sensitivity to Backward Degradation.}}
\vspace{-2mm}
\fontsize{7pt}{9pt}\selectfont
\label{tab:pre-downstream_backward}
\begin{tabular}[t]{c|ccc}
\toprule
\rowcolor{mygray}
 & \multicolumn{3}{c}{\bf Downstream Task Metric} \\
\cline{2-4}
\rowcolor{mygray}
\multirow{-2}{*}{\makecell[c]{\bf Mask \\ \bf Ratio}} & 
\multirow{-1}{*}{\bf minADE$\downarrow$} &
\multirow{-1}{*}{\bf minFDE$\downarrow$} &
\multirow{-1}{*}{\bf MR$\downarrow$} \\
\midrule
 - & 0.3856 & 0.7919 & 0.0854 \\
\midrule

%5\% &  &  &  \\
10\% & 0.3840 \newdrop{0.41\%} & 0.7884 \newdrop{0.44\%} & 0.0840 \newdrop{1.64\%} \\
%15\% &  &  &  \\
%25\% &  &  &  \\
30\% & 0.3860 \newgain{0.10\%}  & 0.7901 \newdrop{0.23\%} & 0.0845 \newdrop{1.05\%} \\
%40\% &  &  &  \\
50\% & 0.3859 \newgain{0.08\%}  & 0.7913 \newdrop{0.08\%} & 0.0834 \newdrop{2.34\%} \\
%60\% &  &  &  \\
70\% & 0.3870 \newgain{0.36\%} & 0.7932 \newgain{0.16\%} & 0.0807 \newdrop{5.50\%} \\
%80\% &  &  &  \\
%90\% &  &  &  \\
100\% & 0.3882 \newgain{0.67\%} & 0.7971 \newgain{0.66\%} & 0.0816 \newdrop{4.45\%} \\

\bottomrule
\end{tabular}
%\vspace{-0.2in}
\end{table}

\noindent\textbf{Experimental Setup.} We adopt general MapTR~\cite{liao2022maptr} as our baseline. In its vectorized output maps, we systematically apply masking to the front and rear regions along the longitudinal axis of the ego vehicle, simulating controlled degradation in these spatial domains. Different masking ratios are applied to either forward or backward areas to emulate varying levels of map quality loss. The perturbed maps are then processed by a standard trajectory prediction model, HiVT~\cite{zhou2022hivt}, and evaluated using established metrics: Minimum Average Displacement Error~(minADE), Minimum Final Displacement Error~(minFDE), and Miss Rate~(MR).

\noindent\textbf{Experimental Analysis.} As shown in Table~\ref{tab:pre-downstream_forward}, Table~\ref{tab:pre-downstream_backward}, and Fig.~\ref{fig:downstream_bias}, our experimental results reveal a distinct asymmetry in how downstream tasks utilize online mapping information. Forward map regions are critical. Masking the forward area leads to a significant and monotonic degradation in all trajectory prediction metrics. When 100\% of the forward region is masked, minADE increases by 9.73\%, and minFDE increases by 13.95\%. This strong correlation confirms that the quality of the forward map is a primary determinant of downstream performance. Backward map regions exert minimal influence on performance. In contrast, masking the backward region results in negligible or slightly positive changes; for instance, a 70\% mask improves the Miss Rate (MR) from 0.0854 to 0.0807. This suggests that the high-fidelity details captured in these regions provide diminishing returns for future-oriented tasks, and that excessive precision can introduce noise, likely due to overfitting to irrelevant contextual details.

\section{Methodology}
\label{sec:methods}

\subsection{Ahead-Aware Metric: A-mAP}
\label{subsec:3-1-A-mAP}

Based on the analysis in Sec.~\ref{sec:pre-exp}, we observe a significant discrepancy in how map information from the forward and backward domains influences downstream task performance. However, most existing online mapping models focus primarily on overall map accuracy, lacking a dedicated metric to assess the precision of forward-facing map regions. To address this gap, we propose Ahead mAP~(A‑mAP), a downstream task‑oriented evaluation metric that quantifies map construction quality specifically in the area ahead of the ego vehicle. Correspondingly, we introduce Rear mAP~(R‑mAP) to evaluate performance in the rear domain. Together, these metrics enable a clearer comparison of forward–backward performance bias across different HD map construction methods.

\noindent\textbf{Metric Computation.} The region of interest (ROI) is divided into two halves relative to the ego vehicle's heading: the area ahead and the rear area. The map construction quality metrics A-mAP and R-mAP are designed to evaluate the matching degree between predicted map elements and ground-truth elements within the ahead and rear regions, respectively. The computation process inside each half-region follows the same procedure as the global mAP, consisting of five stages: spatial alignment, vector matching, true positive identification, precision-recall calculation, and average precision computation. The detailed workflow is provided in the appendix.

\subsection{AMap Framework}
\label{subsec:3-2-framework}

\begin{figure}[t]
\centering
\setlength{\abovecaptionskip}{0.05cm}
\setlength{\belowcaptionskip}{-0.1cm}
\includegraphics[width=\linewidth]{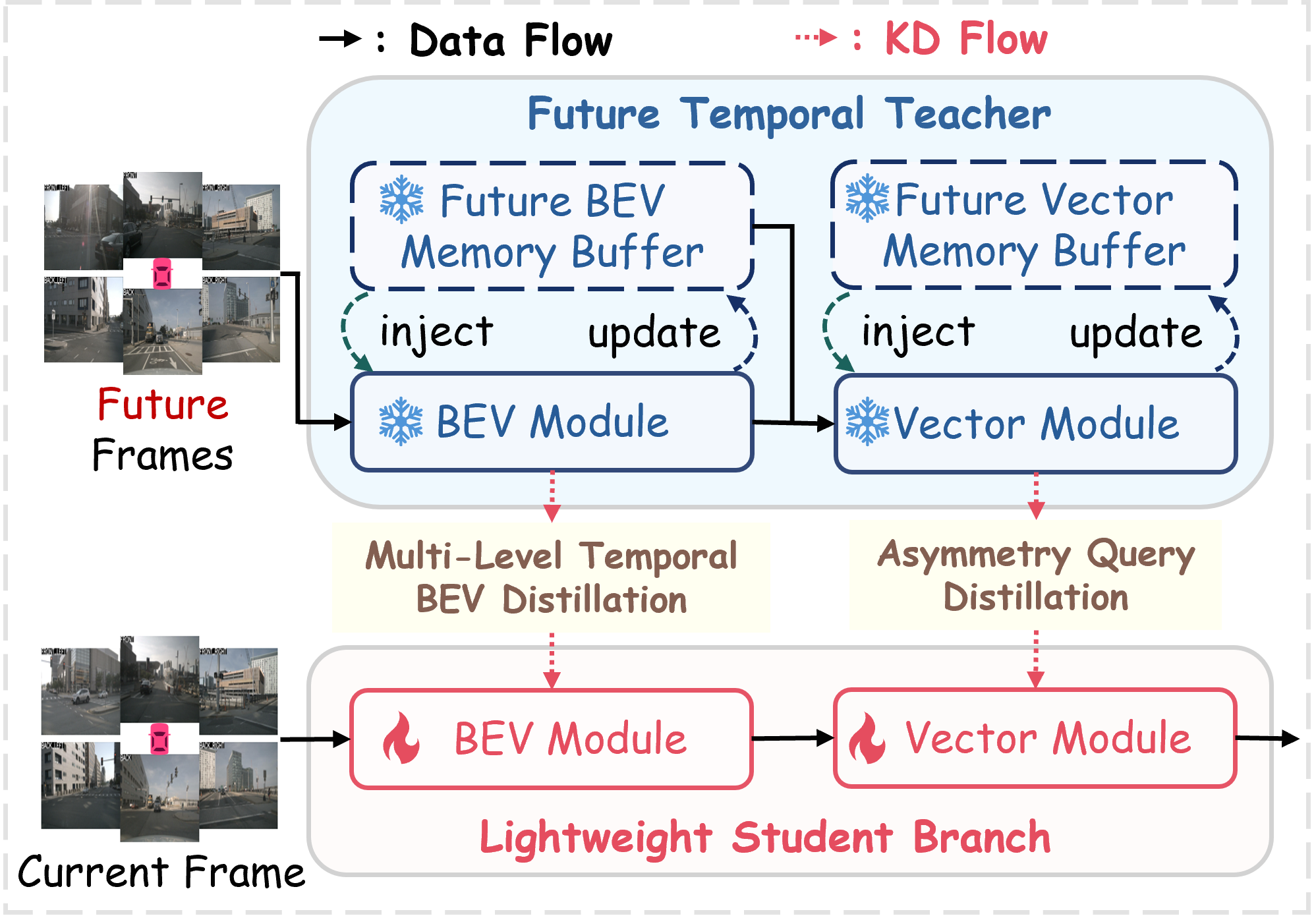}
\caption{\textbf{The overview of our proposed AMap framework.} }
\label{fig:framework}
\vspace{-0.2in}
\end{figure}

\subsubsection{Model Overview}
AMap is designed as a plug-and-play framework to bridge the performance gain bias in existing advanced temporal mapping models. It consists of a teacher model enriched with future temporal information and a lightweight student model, as illustrated in Figure~\ref{fig:framework}. We elaborate the AMap framework in detail using a teacher-student model pair based on the MapTracker \cite{chen2024maptracker} codebase. Notably, the teacher model infused with future temporal information carries no physical significance during inference, since autonomous vehicles cannot access real future sensor inputs during operation. Instead, it serves solely as a privileged information carrier embedding forward-looking knowledge. The student model is a simplified current-frame variant of MapTracker, where temporal information is implicitly infused through knowledge distillation. Consequently, the student model acquires temporal awareness implicitly without introducing additional temporal or spatial overhead during inference. Note that AMap framework is also compatible with teacher models rich in historical information and can be coupled with various student architectures. To facilitate reader comprehension, ``t" denotes the current frame, the subscript ``T" indicates tensors of the teacher model, and the subscript ``S" represents tensors of the student model.
% The teacher model infused with future temporal information is modified from MapTracker~\cite{chen2024maptracker} with historical frame fusion. 
\noindent\textbf{Teacher with Future Priors.} Our teacher model comprises 4 key components: the BEV Module, Future BEV Memory Buffer, Vector Module, and Future Vector Memory Buffer. Image features are extracted from multi-view RGB images through an image backbone. The BEV Module transforms the future BEV ``memory" $B^{t+1}$ to the current frame's coordinate system based on pose information, after which the basic BEV features $B^{t,basic}_{T}$ of the current frame are obtained using spatial-deformable cross-attention similar to BEVFormer~\cite{li2024bevformer}. Furthermore, the 4 future ``cached" BEV features from the Future BEV Memory Buffer are fused with the current features to derive refined BEV features $B^{t,refined}_{T}$. The refined BEV features are injected into the Future BEV Buffer while being passed to the Vector Module. The Vector Module first retrieves the query ``memory" $Q_{T}^{t+1}$ for the future $t+1$ frame from the Future Vector Memory Buffer and aligns it to the current frame via an MLP. The current frame's queries are initialized as:
\begin{equation}
\vspace{-1ex}
Q_T^{t,basic}=[Q_T^{t+1_{prop}},Q_T^{t,new}],
\end{equation}
where $Q_T^{t+1_{prop}}$ denotes ``tracking" elements from the $t+1$ frame and $Q_T^{t,new}$ represents 100 new candidate queries. Subsequently, self-attention and BEV feature-based cross-attention are applied to enhance contextual representation. For each ``tracking" element in the current frame, its corresponding future latent vector is selected from the Future Vector Memory Buffer for query fusion, yielding a refined query $Q_T^{t,refined}$. Finally, the filtered refined queries are fed into the detection head to produce the mapping results.

\noindent\textbf{Lightweight Student.} The lightweight current-frame student model is built upon the original MapTracker~\cite{chen2024maptracker}, with temporal fusion modules removed from both the BEV and Query dimensions. The base BEV features of the student branch $B_S^{t,basic}$ are obtained by transforming 2D image features, after which refined BEV features $B_S^{t,refined}$ are acquired through contextual enhancement. Due to the absence of a query buffer, the initialized learnable queries $Q_S^{t,new}$ are optimized via self-attention and BEV feature-based cross-attention. Similarly, prediction results of map elements are generated based on filtered high-confidence queries.

\begin{table*}[htbp]
\setlength{\tabcolsep}{3.0pt}
\setlength{\abovecaptionskip}{0.1cm}
\centering
\caption{\textbf{Quantitative comparisons with non-KD camera-based online HD mapping models on the nuScenes validation set.}}
\fontsize{8pt}{10pt}\selectfont
\label{tab:main_table}
\begin{tabular}[t]{lcc|cccc|cccc|c|cc}
\toprule
\rowcolor{mygray}
\bf Method & 
\bf Temporal &
\bf Backbone &
\bf AP$_{p.}\uparrow$ &
\bf AP$_{d.}\uparrow$ &
\bf AP$_{b.}\uparrow$ &
\bf mAP$\uparrow$ &
\bf A-AP$_{p.}\uparrow$ &
\bf A-AP$_{d.}\uparrow$ &
\bf A-AP$_{b.}\uparrow$ &

\bf A-mAP$\uparrow$ &
\bf R-mAP$\uparrow$ &
\bf FPS$\uparrow$ 
\\
\midrule
MapTR~(GKT) & \ding{51} & R50 & 45.99 & 53.39 & 54.46 & 51.28 & 43.64 & 54.89 & 58.14 & 52.23 & 55.22 & 14.7 \\
MapTR~(BEVFormer) & \ding{51} & R50 & 51.71 & 54.88 & 53.29 & 53.29 & 48.80 & 56.86 & 55.41 & 53.69 & 55.18 & 15.6 \\
StreamMapNet &  \ding{51} & R50 & 48.36 & 52.44 & 52.65 & 51.15 & 49.44 & 57.5 & 57.89 & 54.94 & 50.94 & 15.7\\
HRMapNet & \ding{51} & R50 & 62.46 & 64.81 & 66.33 & 64.53 & 57.63 & 63.79 & 65.61 & 62.34 & 69.51 & 16.2 \\
MapTracker  & \ding{51} & R18 & 71.68 & 67.25 & 69.09 & 69.34 & 64.06 & \textbf{68.75} & 67.76 & 66.86 & 76.51 & \textbf{22.3}\\
MapTracker  & \ding{51} & R50  & \textbf{72.73} & \textbf{75.77} & \textbf{70.31} & \textbf{72.93} & \textbf{74.66} & 65.75 & \textbf{69.69} & \textbf{70.03} & \textbf{78.92} & 15.6  \\
\midrule
MapTR  & \ding{55} & R50  & 36.18 & 48.44 & 47.67 & 44.10 & 35.82 & 49.48 & 49.73 & 45.01 & 48.87 & 17.2\\
GeMap  & \ding{55} & R50 & 45.62 & 53.95 & 54.20 & 51.26 & 41.37 & 56.91 & 56.56 & 51.61 & 54.74 & 13.7 \\
MGMap  & \ding{55} & R50 & 48.21 & 55.80 & 55.43 & 53.15 & 48.56 & 57.21 & 59.64 & 55.13 & 56.78 & 12.4 \\
MapTRv2 & \ding{55} & R50 & 60.18 & 61.33 & 62.62 & 61.38 & 60.42 & 64.78 & 67.44 & 64.21 & 61.82 & 14.4\\
MapQR & \ding{55} & R50  & 63.36 & 68.03 & \textbf{67.67} & 66.35 & 62.47 & 71.20 & \textbf{71.47} & 68.38 & 65.59 & 12.4 \\
MapTracker & \ding{55} & R50  & 68.21 & 69.71 & 67.00 & 68.30 & 64.33 & 73.25 & 69.83 & 69.30 & 69.47 & \textbf{20.1} \\
\textbf{AMap~(Ours)} & \ding{55} & R50  & \textbf{69.97}  & \textbf{70.13} & \textbf{67.67} & \textbf{69.26} & \textbf{65.09} & \textbf{74.48} & 71.01 & \textbf{70.19} & \textbf{69.61} & \textbf{20.1} \\
\midrule
MapTR & \ding{55} & R18  & 26.06 & 33.68 & 37.32 & 32.35 & 29.84 & 37.06 & 42.11 & 36.34 & 32.76 & 31.4  \\
MapTRv2 & \ding{55} & R18  & 54.07 & 59.12 & 58.25 & 57.15 & 56.69 & 62.10 & 63.65 & 60.81 & 56.17 & 16.6 \\
MapQR & \ding{55} & R18 & 59.33 & 63.26 & \textbf{64.35} & 62.31 & 60.03 & 66.09 & \textbf{67.98} & 64.70 & 63.13 & 13.7 \\
MapTracker & \ding{55} & R18  & 61.01 & 65.28 & 62.14 & 62.81 & 59.33 & 69.33 & 65.23 & 64.63 & 64.04 & \textbf{31.5} \\
\textbf{AMap~(Ours)} & \ding{55} & R18  & \textbf{63.11} & \textbf{67.08} & 63.27 & \textbf{64.49} & \textbf{60.70} & \textbf{71.96} & 66.16 & \textbf{66.28} & \textbf{65.11} & \textbf{31.5} \\
\bottomrule
\end{tabular}

\begin{tablenotes}
\item[1] \scriptsize{\textbullet\; AP$_{p.}$, AP$_{d.}$, AP$_{b.}$ denote the Average Precision (AP) for pedestrian crossings, lane dividers, and road boundaries, respectively.}
\end{tablenotes}

\vspace{-0.1in}
\end{table*}

\subsubsection{AMap Components}

\noindent\textbf{Multi-Level Temporal BEV Distillation.} Following the paradigms established by BEVDistill~\cite{chen2022bevdistill} and MapDistill~\cite{hao2024mapdistill}, we leverage knowledge distillation on BEV embeddings to transfer spatial-semantic information. However, distinct from previous approaches that primarily target the initial BEV features, we extend the distillation scope to the \textit{refined} BEV representations, which are subsequently employed for the auxiliary segmentation task. We identify that applying global distillation directly to the entire feature map inevitably introduces background noise and leads to optimization imbalance due to the sparsity of foreground semantics. To mitigate this, we propose a ground-truth-guided masking strategy. Specifically, we utilize the segmentation ground truth to generate a spatial mask $\mathcal{M}$, which suppresses irrelevant background signals and focuses the distillation on informative regions. The masked feature distillation loss is formulated as:
\begin{equation}
\vspace{-1ex}
\small
    \mathcal{L}_{feat} = \frac{1}{\sum_{i,j} \mathcal{M}_{i,j}} \sum_{i=1}^{H} \sum_{j=1}^{W} \mathcal{M}_{i,j} \left\| \mathbf{F}_{\text{S}}^{(i,j)} - \mathbf{F}_{\text{T}}^{(i,j)} \right\|_2,
    \label{eq:feat_distill}
\end{equation}
\noindent where $\mathbf{F}_{\text{T}} \in \mathbb{R}^{C \times H \times W}$ and $\mathbf{F}_{\text{S}} \in \mathbb{R}^{C \times H \times W}$ denote the basic and refined BEV feature maps from the teacher and student branches, respectively. $\mathcal{M}_{i,j} \in \{0, 1\}$ represents the binary mask derived from the segmentation ground truth, serving to spatially reweight the distillation loss.

\noindent\textbf{Asymmetry Query Distillation.} Distilling knowledge from time-series models that utilize dynamically tracked queries presents unique challenges, rendering conventional distillation techniques ill-suited. Methods demanding direct query-to-query alignment, such as MapDistill~\cite{hao2024mapdistill}, are not directly applicable by this dynamic context. Moreover, our preliminary experiments confirmed that a naive approach is inherently limited in distilling a fixed set of static queries. To overcome this limitation, inspired by DETRDistill~\cite{chang2023detrdistill}, we employ a matching-based dynamic query distillation strategy. This approach specifically addresses the problem of temporal query inconsistency inherent to models like MapTracker~\cite{chen2024maptracker}.

In detail, We first find one-to-one matching $\hat{\sigma}$ from the student's $N_{\text{S}}$ queries (e.g., 100) to the teacher's $N_{\text{T}}$ queries (e.g., 116) using the Hungarian algorithm.
The logits distillation loss $\mathcal{L}_{\text{logitsKD}}$ is then computed only on these matched pairs:
\begin{equation}
\vspace{-2ex}
    \mathcal{L}_{\text{logitsKD}} = \sum_{i=1}^{N_{\text{S}}} \mathcal{L}_{\text{KL}}\left( \text{Logits}_{\text{S}}[i] \parallel \text{Logits}_{\text{T}}[\hat{\sigma}(i)] \right)
\end{equation}
This formula sums the Kullback-Leibler divergence ($\mathcal{L}_{\text{KL}}$) over all $N_{\text{S}}$ student queries. For each student query $i$, we distill its output $\text{Logits}_{\text{S}}[i]$ against the logits of the teacher query it was matched with, $\text{Logits}_{\text{T}}[\hat{\sigma}(i)]$. The mapping $\hat{\sigma}(i)$ provides the index of the specific teacher query that was optimally assigned to student query $i$.

\definecolor{demogray}{gray}{0.95}
% Define header gray if not already defined
% \definecolor{mygray}{gray}{0.9}

\begin{table*}[htbp]
\setlength{\tabcolsep}{4.5pt}
\setlength{\abovecaptionskip}{0.1cm}
\centering
\caption{\textbf{Quantitative comparisons with Other KD methods on the nuScenes validation set.}}
\fontsize{8pt}{10pt}\selectfont
\label{tab:main_table_withKD}
\begin{tabular}[t]{l|c|c|cc|cccc|cc}
\toprule
\rowcolor{mygray}
\bf Distillation Setup & % Renamed from Group ID
\bf Setting &
\bf Method & 
\bf Temporal &
\bf Backbone &
\bf AP$_{p.}\uparrow$ &
\bf AP$_{d.}\uparrow$ &
\bf AP$_{b.}\uparrow$ &
\bf mAP$\uparrow$ &
\bf A-mAP$\uparrow$ &
\bf R-mAP$\uparrow$ 
\\
\midrule

% --- GROUP 1: Future to Static (White Background for Emphasis) ---
\multirow{5}{*}{\textbf{Future $\to$ Static}} & % Renamed for clarity
T & MapTracker & 5 Future & R50 & 77.00 & 75.65 & 75.53 & 76.06 & 79.86 & 73.32 \\
& S & MapTracker & None & R18 & 61.01 & 65.28 & 62.14 & 62.81 & 64.63 & 64.04 \\
 & KD &  BEVDistill* & None & R18 & 56.33 & 60.75 & 56.82 & 57.97 \drop{4.84} & 60.08 \drop{4.55} & 60.06 \\
 & KD &  MapDistill* & None & R18 & 57.08 & 59.60 & 56.78 & 57.82 \drop{5.61} & 59.97 \drop{4.66} & 59.87 \\
 & KD & \textbf{AMap (Ours)} & None & R18 & \textbf{63.11} & \textbf{67.08} & \textbf{63.27} & \textbf{64.49} \gain{1.68} & \textbf{66.28} \gain{1.65} & \textbf{65.11} \\

\midrule

% --- GROUP 2: Past to Static (Light Gray) ---
\rowcolor{demogray}
 & T & MapTracker & 5 Past & R50 & 72.73 & 75.77 & 70.31 & 72.93 & 70.03 & 78.92 \\
\rowcolor{demogray}
 & S & MapTracker & None & R18 & 61.01 & 65.28 & 62.14 & 62.81 & 64.63 & 64.04 \\
\rowcolor{demogray}
 & KD & BEVDistill* & None & R18 & 54.48 & 58.47 & 55.49 & 56.15 \drop{6.66} & 58.47 & 58.86  \\
\rowcolor{demogray}
 & KD & MapDistill* & None & R18 & 56.03 & 59.53 & 57.32 & 57.63 \drop{5.18}& 59.72 & 59.83  \\
\rowcolor{demogray}
\multirow{-5}{*}{\textbf{Past $\to$ Static}} & KD & \textbf{AMap (Ours)} &  None & R18 & \textbf{62.72} & \textbf{65.92} & \textbf{62.26} & \textbf{63.63} \gain{0.82} & \textbf{65.08} & \textbf{65.08} \\
% Note: moved multirow to the bottom with negative count (-5) to ensure it renders on top of the color

\midrule

% --- GROUP 3: Past to Past (Light Gray) ---
\rowcolor{demogray}
 & T & MapTracker & 5 Past & R50 & 72.73 & 75.77 & 70.31 & 72.93 & 70.03 & 78.92 \\
\rowcolor{demogray}
 & S & MapTracker & 5 Past & R18 & 71.68 & 67.25 & 69.09 & 69.34 & \textbf{66.86} & 76.51 \\
\rowcolor{demogray}
 & KD & BEVDistill* & 5 Past & R18 & 70.29 & 68.95 & 68.70 & 69.31 \drop{0.03} & 65.91 & 76.69 \\
\rowcolor{demogray}
 & KD &  MapDistill* & 5 Past & R18 & 67.85 & 67.61 & 67.77 & 67.74 \drop{1.6} & 64.75 & 75.00 \\
\rowcolor{demogray}
\multirow{-5}{*}{\textbf{Past $\to$ Past}} & KD & \textbf{AMap (Ours)} & 5 Past & R18 & \textbf{71.65} & \textbf{68.96} & \textbf{70.51} & \textbf{70.37} \gain{1.03} & 65.99 & \textbf{79.95} \\

\bottomrule
\end{tabular}
\begin{tablenotes}
\item[1] \scriptsize{\textbullet\; The table compares different temporal distillation settings.}
\item[2] \scriptsize{\textbullet\; \textbf{Future $\to$ Static}: Distilling future frames into a static student. \textbf{Past $\to$ Static}: Distilling history into a static student.}
\item[3] \scriptsize{\textbullet\; *: Temporal distillation completion version.}
\end{tablenotes}
\vspace{-0.2in}
\end{table*}

\section{Experiments}
\label{sec:exp}

\subsection{Experimental Settings}

\noindent\textbf{Dataset.}
The nuScenes \cite{caesar2020nuscenes} and Argoverse 2 \cite{wilson2023argoverse} datasets are used to comprehensively evaluate the effectiveness of our approach. The nuScenes dataset contains 1,000 scene sequences, with keyframes annotated at 2 Hz. Each scene sequence consists of a 20-second video clip captured by 6 synchronized surrounding cameras. The Argoverse 2 dataset is recorded by 7 surrounding cameras and annotated at 10 Hz. The scene partitioning rules and the ground truth generation strategy follow previous work \cite{chen2024maptracker}.

\noindent\textbf{Evaluation Metric.} Both the classic global metric mAP and our proposed A-mAP and R-mAP are employed to comprehensively evaluate previous works and our AMap framework. Following established protocols, we consider three map element categories: pedestrian crossing, lane divider, and road boundary.

\noindent\textbf{Training.}
The training consists of two stages: teacher model pre-training and joint teacher-student training. During the joint teacher-student training phase, the teacher model parameters loaded from the pre-trained weights are entirely frozen, and only the student model parameters are updated. For the teacher model, we follow the default settings of MapTracker, where the temporal sampling strategy involves randomly selecting 4 frames from 10 consecutive frames in chronological order. We use the AdamW \cite{loshchilov2017adamW} optimizer with an initial learning rate of 5e-4 and weight decay of 0.01. A cosine learning rate scheduler is employed, decaying to a final learning rate of 1.5e-6. All models are trained on 8 NVIDIA H20 GPUs.

% We build our distillation framework based on the MapTracker \cite{chen2024maptracker} codebase.  For the teacher model, we follow the default settings of MapTracker, where the temporal sampling strategy involves randomly selecting 4 frames from 10 consecutive frames in chronological order.

\subsection{Comparisons with State-of-the-art Methods}
\noindent\textbf{Benchmark and Analysis.} To comprehensively evaluate the performance of existing online HD map construction methods in the front and rear spatial domains, we conduct a systematic assessment of current state-of-the-art camera-based methods on the nuScenes~\cite{caesar2020nuscenes} validation set. As shown in Table~\ref{tab:main_table}, temporal models \cite{liao2022maptr,yuan2024streammapnet,zhang2024hrmapnet,chen2024maptracker} demonstrate a clear advantage in global performance metrics. However, most of them exhibit a notable asymmetry between the gains in A-mAP and R‑mAP. For instance, MapTracker shows a large performance gap exceeding 8 points under both ResNet-18 and ResNet-50 backbones. This bias suggests that the performance gains of most existing temporal models primarily stem from improvements in the rear region, while a performance bottleneck remains in the forward area, which is more critical for downstream tasks.

In contrast, current-frame models \cite{liao2022maptr,zhang2024gemap,liu2024mgmap,liao2025maptrv2,liu2024mapqr} achieve a more balanced improvement between forward and rear metrics. Our proposed AMap distillation framework is designed to transfer forward-aware knowledge, embedded in a teacher model with future temporal priors, to a lightweight current-frame student model via knowledge distillation. Experimental results in Table~\ref{tab:main_table} show that the AMap framework effectively improves mAP, A‑mAP, and R‑mAP without introducing extra inference cost. It is worth emphasizing that, unlike other temporal models that tend to boost rear region performance, the implicit injection of future temporal priors in AMap leads to a distinct gain in forward spatial perception.

\noindent\textbf{Comparison with Other Knowledge Distillation Methods.} Previous knowledge distillation approaches in autonomous driving have predominantly focused on cross-modal knowledge transfer. As summarized in Table~\ref{tab:main_table_withKD}, we design 3 comparative distillation settings: (1) distillation from a teacher model with future temporal information to a current-frame student model; (2) distillation from a teacher with past temporal information to a current-frame student; and (3) distillation from a teacher with past temporal inputs to a lightweight student that also incorporates historical information.

These 3 configurations demonstrate the superiority of our AMap distillation framework in temporally-aware distillation. Existing methods such as BEVDistill~\cite{chen2022bevdistill}, and MapDistill~\cite{hao2024mapdistill}, are generally designed for current-frame to current-frame knowledge transfer, making them less suitable for distilling temporal knowledge into a current-frame model. The stronger performance observed in the third group further confirms that distillation is more effective when the teacher and student operate under aligned temporal receptive fields, as compared to the asymmetric settings in groups ``Future $\to$ Static" and ``Past $\to$ Static".

\begin{table}[htbp]
\setlength{\abovecaptionskip}{0.1cm}
\centering
\caption{\textbf{Results based on the MapTR model.}}
\fontsize{7pt}{9pt}\selectfont
\label{tab:ablation_maptr}
\begin{tabular}[t]{lccccc}
\toprule
\rowcolor{mygray}
\bf Setting & 
\bf Tem. &
\bf Back. & 
\bf mAP$\uparrow$ &
\bf A-mAP$\uparrow$ &
\bf R-mAP$\uparrow$ 
\\
\midrule
Teacher & 5 Future & R50 & 51.85 & 57.01 & 50.12 \\
\midrule

Baseline & \ding{55} & R18 & 32.35 & 36.34 & 32.76 \\
+ Ours & \ding{55} & R18 & 47.84 & 51.29\gain{14.95}  & 47.59 \\
\bottomrule
\end{tabular}
\vspace{-0.1in}
\end{table}

\begin{table}[htbp]
\setlength{\abovecaptionskip}{0.1cm}
\centering
% \caption{\textbf{Quantitative comparisons with non-KD camera-based online HD mapping models on the Argoverse2 validation set.}}
\caption{\textbf{Results on the Argoverse2 validation set.}}
\fontsize{7pt}{9pt}\selectfont
\label{tab:main_table_av2}
\begin{tabular}[t]{lccccc}
\toprule
\rowcolor{mygray}
\bf Method & 
\bf Tem. &
\bf Back. & 
\bf mAP$\uparrow$ &
\bf A-mAP$\uparrow$ &
\bf R-mAP$\uparrow$ 
\\
\midrule
%MapTracker & 5 & R18 & \todo{TODO} & \todo{TODO} & \todo{TODO} \\
MapTracker & 5 past & R50 & 76.87 & 73.00 & 83.82 \\
MapTracker & 5 future & R50 & 75.57 & 80.79 & 72.42
\\
\midrule
% MapTR & None & R50 & 57.5* & \todo{TODO} & \todo{TODO} \\
% MapTRv2 & None & R50 & 67.4* & \todo{TODO} & \todo{TODO} \\
MapTracker & None & R18 & 63.69 & 65.29 & 66.10 \\
+ Ours & None & R18 & \textbf{65.25} & \textbf{67.81} & \textbf{67.53} \\

\bottomrule
\end{tabular}
\vspace{-0.2in}
\end{table}

\subsection{Experiments on Generalization}

\noindent\textbf{Experiments on the MapTR model.} The proposed AMap distillation framework is integrated into the MapTR~\cite{liao2022maptr} architecture. The teacher model is built upon a ResNet-50 backbone and incorporates future temporal information, while the student model employs a lightweight ResNet-18 backbone and uses only current-frame input. As shown in Table~\ref{tab:ablation_maptr}, experimental results demonstrate that the student model equipped with the AMap distillation strategy achieves significant performance improvements compared to the baseline student model, even approaching the performance level of the teacher model. This validates the strong plug-and-play capability of the AMap framework.

\noindent\textbf{Evaluation on Argoverse 2 Dataset.} As shown in Table \ref{tab:main_table_av2}, the AMap distillation framework also demonstrates strong performance on the Argoverse 2 dataset~\cite{wilson2023argoverse}. Compared to the baseline student model, the student model equipped with our AMap distillation strategy achieves consistent improvements across all three evaluation metrics, including mAP, A-mAP and R-mAP. Particularly noteworthy is the significant performance gain of 2.52 A-mAP observed in the forward perception region.

\begin{table}[htbp]
\setlength{\abovecaptionskip}{0.1cm}
\centering
\caption{\textbf{Effect of key designs in AMap.}}
\fontsize{7pt}{9pt}\selectfont
\label{tab:ablation_main}
\begin{tabular}[t]{cc|c|ccc}
\toprule

% ===== first =====
\rowcolor{mygray}

\multicolumn{2}{c|}{\bf BEV Level}
& & & & \\
\cline{1-2}
% ===== second =====
\rowcolor{mygray}
\multirow{-1}{*}{\makecell[c]{\bf Basic}} &
\multirow{-1}{*}{\makecell[c]{\bf Refined }} &
\multirow{-2}{*}{\makecell[c]{\bf Query \\ \bf Level }} &
\multirow{-2}{*}{\bf mAP$\uparrow$} &
\multirow{-2}{*}{\bf A-mAP$\uparrow$ } &
\multirow{-2}{*}{\bf R-mAP$\uparrow$ } 
\\

\midrule

\ding{55} & \ding{55} & \ding{55} & 62.81 & 64.63 & 64.04 \\
\midrule
\ding{51} & \ding{55} & \ding{55} & 62.66 & 64.33 & 64.31 \\
\ding{55} & \ding{51} & \ding{55} & 62.72 & 64.68 & 64.78 \\
\ding{51} & \ding{51} & \ding{55} & 63.47 & 65.20 & \textbf{65.54} \\
\ding{55} & \ding{55} & \ding{51} & 63.57 & 64.61 & 64.91 \\
\ding{51} & \ding{51} & \ding{51} & \textbf{64.49} & \textbf{66.28} & 65.11 \\
\bottomrule
\end{tabular}
\vspace{-0.1in}
\end{table}
\begin{table}[htbp]
\setlength{\abovecaptionskip}{0.1cm}
\centering
\caption{\textbf{Impact of Asymmetry Query Distillation.}}
\fontsize{7pt}{9pt}\selectfont
\label{tab:ablation_query}
\begin{tabular}[t]{l|cccc}
\toprule
\rowcolor{mygray}
\bf Method & 
\bf AP$_{p.}\uparrow$ &
\bf AP$_{d.}\uparrow$ &
\bf AP$_{b.}\uparrow$ &
\bf mAP$\uparrow$ 
\\
\midrule
baseline & 61.01 & 65.28 & 62.14 & 62.81 \\
+ BEV Level & 62.49 & 65.18 & 62.09 & 63.25 \\
\midrule
+ MapDistill (dummy)  & 56.03 & 59.53 & 57.32 & 57.63 \\ %\drop{5.62} \\
+ MapDistill (Top K)  & 53.55 & 58.86 & 57.42 & 56.61 \\ %\drop{6.64} \\
+ Hungrian (MSE)  & 62.46 & 64.24 & \textbf{62.26} & 62.99 \\ % \drop{0.26} \\
+ Hungrian (Cosine)  & 60.99 & 65.02 & 61.27 & 62.43 \\ % \drop{0.82} \\
% + AutoMatch  & 60.96 & 65.69 & 61.79 & 62.81 \\ % \drop{0.44} \\
+ Ours & \textbf{62.72} & \textbf{65.92} & \textbf{62.26} & \textbf{63.63} \\ %\gain{0.38} \\
\bottomrule
\end{tabular}
% \begin{tablenotes}
% \item[1] \scriptsize{\textit{\textbullet\; Gains are calculated based on the baseline with BEV level distillation.}}
% \end{tablenotes}
\vspace{-0.2in}
\end{table}

\begin{figure*}[t]
    \centering
    \includegraphics[width=\linewidth]{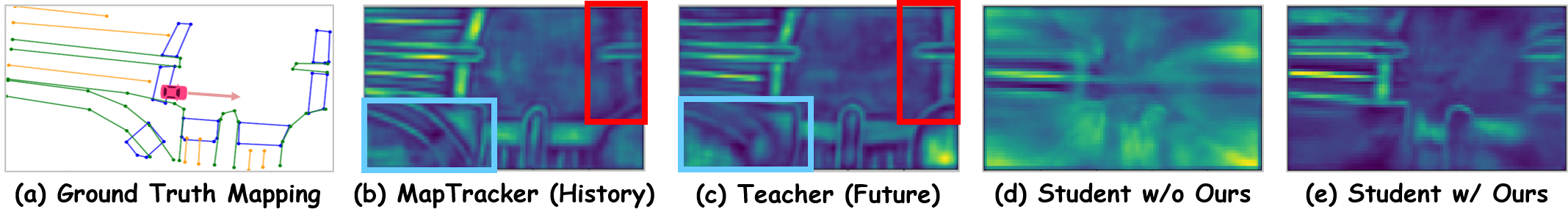}
    \vspace{-6mm}
    \caption{BEV feature visualization comparison among: (b) historical sequence-based MapTracker, (c) our teacher model with future priors, (d) the baseline student model, and (e) the student model with our proposed distillation strategy. \textcolor{fig3_blue}{Blue} boxes and \textcolor{red}{red} boxes indicate the advantages of historical sequences and future sequences, respectively.}
    \label{fig:bev}
    \vspace{-5mm}
\end{figure*}

\subsection{Ablation Study}
\noindent\textbf{Ablation study of core components in AMap.} As shown in Table~\ref{tab:ablation_main}, we analyze our proposed BEV-level and query-level distillation strategies through different combinations. Comparing the first three ablation groups, we observe that employing a single BEV-level distillation strategy has limited impact on map construction accuracy. However, when both the refined and basic BEV features are used as loss terms, the student model achieves noticeable improvement. Furthermore, we find that the two half-region metrics, A‑mAP and R‑mAP, are not strictly correlated with the global mAP. For instance, when only the refined BEV feature is used for supervision, both A‑mAP and R‑mAP increase to varying degrees, while the overall mAP decreases. We hypothesize that this phenomenon may be related to the region-splitting strategy; a detailed analysis is provided in the appendix. Additionally, we demonstrate that combining query-level and BEV-level supervision maximizes the distillation gains, yielding the best overall performance.

\noindent\textbf{Ablation Study on Different methods for Asymmetry Query Distillation.} As shown in Table~\ref{tab:ablation_query}, we ablate different distillation matching strategies on this asymmetry query problem. The baseline model achieves 63.25 mAP. Naive matching strategies, such as MapDistill \cite{hao2024mapdistill} applied only on dummy queries or Top-K (K=100) queries in Maptracker \cite{chen2024maptracker}, fail to establish proper student-teacher correspondence and cause a severe performance collapse, dropping to 57.63 and 56.61 mAP, respectively. We then evaluate more advanced assignment methods. 
% A learnable matching approach (+ AutoMatch\TODO{cite?}) shows no improvement over the baseline (62.81 mAP), indicating optimization difficulty. 
Using Hungarian matching for assignment is sensitive to the loss function and fail to get performance gain when distill query embeddings: Hungarian (Cosine) and Hungarian (MSE) both degrade performance. Finally, our method when apply this distillation on logits using KL loss achieving the highest scores overall mAP (63.63).

% \input{tables/7_downstream}
% \subsection{Zero-Shot Downstream Task Performance}
% \TODO{todotodotodo}

\subsection{Qualitative Results}
\noindent\textbf{Visual Comparison of BEV Features.} As illustrated in Fig.~\ref{fig:bev}, the teacher model with future temporal priors and the historically-driven MapTracker exhibit complementary strengths in scene representation: MapTracker~(b) demonstrates stronger representational capacity in the rear spatial region, while the teacher model~(c) achieves more discriminative scene perception ahead of the ego vehicle. Compared to the baseline student model~(d), the BEV features of our AMap-enhanced student model~(e) show significant improvement.

\begin{figure}[t]
\centering
\setlength{\abovecaptionskip}{0.05cm}
\setlength{\belowcaptionskip}{-0.1cm}
\includegraphics[width=\linewidth]{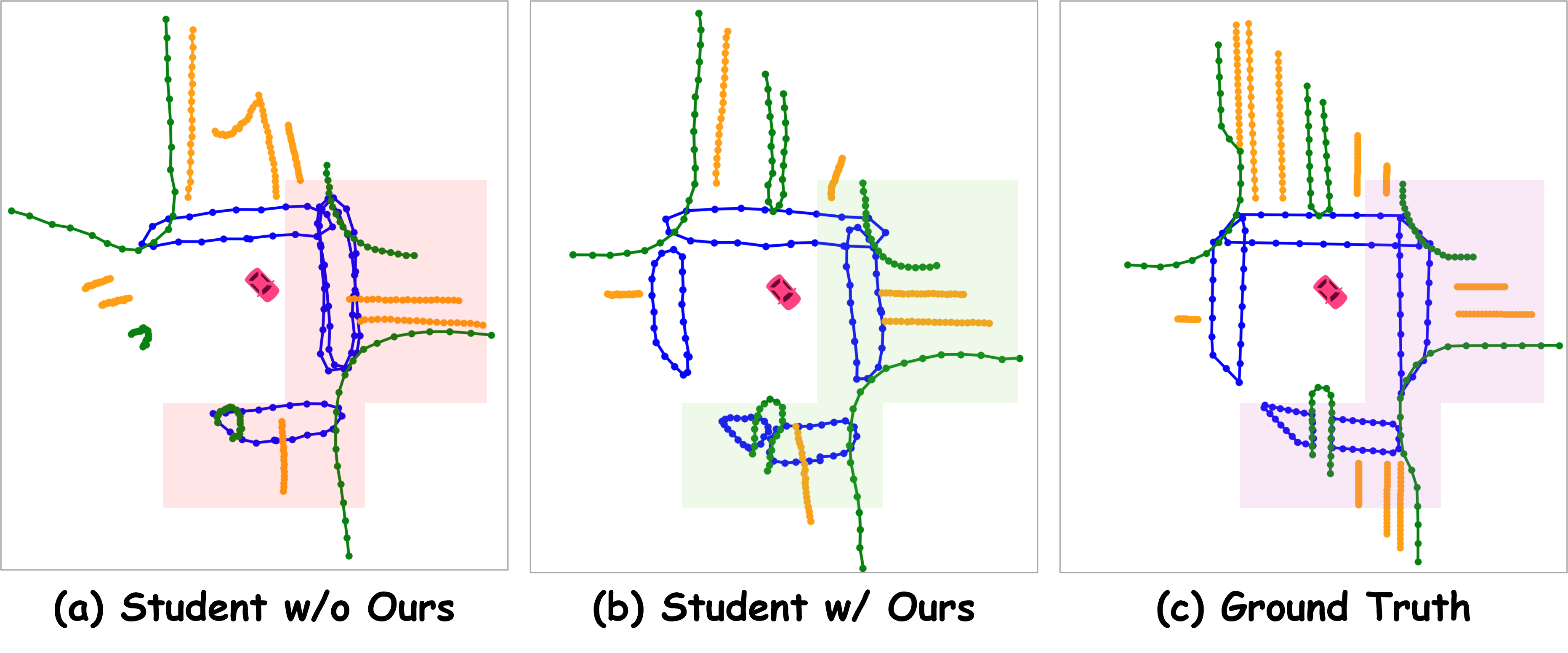}
\caption{\textbf{Qualitative results of the baseline student model~(a), AMap (Ours)~(b), and Ground-Truth~(c).}}
\label{fig:vis}
\vspace{-0.2in}
\end{figure}

\noindent\textbf{Visual Comparison in Complex Intersection Scenarios.} As shown in Fig.~\ref{fig:vis}, the student model enhanced with AMap distillation demonstrates significantly stronger ahead-aware mapping capability compared to the baseline student model. This advantage is particularly evident in its accurate reconstruction of map elements such as pedestrian crossings and lane dividers in the forward driving direction, as well as all three categories of map elements at intersections to the right of the ego vehicle.

% Fig.~\ref{fig:bev}~(b)-(c) reveal complementary advantages of historical and future sequences: historical sequences yield clearer feature activations in rear regions (blue box), while future priors produce sharper representations in front regions (red box), validating our integration approach. Furthermore, Fig.~\ref{fig:bev}~(d)-(e) demonstrate that our distillation strategy significantly enhances the student's BEV features, particularly in front areas (red box), where features become notably sharper and more continuous, successfully transferring the teacher's front-aware knowledge.

\section{Conclusion}
\label{sec:conclusion}

In this work, we have critically re-examined the prevailing temporal fusion paradigm in online HD map reconstruction and identified a severe ``backward-looking" bias. Our investigation into downstream trajectory prediction tasks reveals a counter-intuitive but pivotal insight: high-precision reconstruction of the rearward environment is not only redundant but potentially distracting; notably, masking out rearward perception can surprisingly improve trajectory prediction performance. This highlights a fundamental misalignment between standard global mAP metrics and the decision-centric  priorities of autonomous driving.

To resolve this, we proposed AMap, a novel framework that pioneers a ``distill-from-future" mechanism. By treating future temporal context as privileged information, we successfully transfer ``look-ahead" capabilities to a current-frame student. Our method achieves remarkable results: using a lightweight ResNet-18 backbone, AMap establishes a new state-of-the-art for forward-region perception. Crucially, it outperforms both historical distillation baselines and computationally heavy temporal models in the critical frontal view, even when the latter exhibit higher overall mAP due to their rearward bias. We believe this work charts a new course for the field, encouraging the community to rethink training strategies that prioritize task-relevant foresight over uniform reconstruction for predictive scene understanding.

\section{Acknowledgements}
\label{sec:acknowledgements}

This work was supported in part by the Beijing Natural Science Foundation under Grant L243008, in part by the Shandong Provincial Natural Science Foundation under Grant ZR2024LZN010, in part by the National Natural Science Foundation of China (NSFC) under Grant 52441202, and in part by the National Key R\&D Program of China under Grant 2024YFB43000303.
% {
%     \small
%     \bibliographystyle{ieeenat_fullname}
%     \bibliography{main}
% }

%     \small
%     \bibliographystyle{ieeenat_fullname}
%     \bibliography{main}
% }

\clearpage
\setcounter{page}{1}
\maketitlesupplementary

In the supplementary material, we provide additional details covering preliminary experiments and discussions, experimental results and analysis of the AMap distillation framework, visualizations, training and inference specifics (including computational costs), calculation details of the A-Map metric, and a more comprehensive Ahead-aware Online Mapping benchmark with related analysis. The supplementary material is organized as follows:

\begin{itemize}
\item Sec.~\ref{sup_sec:1_pre_exp} presents additional preliminary experiments and discussions to better illustrate the impact of ahead-aware mapping on downstream tasks.
\item Sec.~\ref{sup_sec:2_ablation_and_vis} includes further ablation studies and more detailed visual comparisons.
\item Sec.~\ref{sup_sec:3_details} describes additional training and inference configurations, together with an analysis of computational costs.
\item Sec.~\ref{sup_sec:4_benchmark} offers an extended A-mAP vs R-mAP benchmark and corresponding analysis on single frame models.
\end{itemize}

\section{Extended Preliminary Experiments}
\label{sup_sec:1_pre_exp}

\begin{figure}[t]
\centering
\setlength{\abovecaptionskip}{0.05cm}
\setlength{\belowcaptionskip}{-0.1cm}
\includegraphics[width=0.90\linewidth]{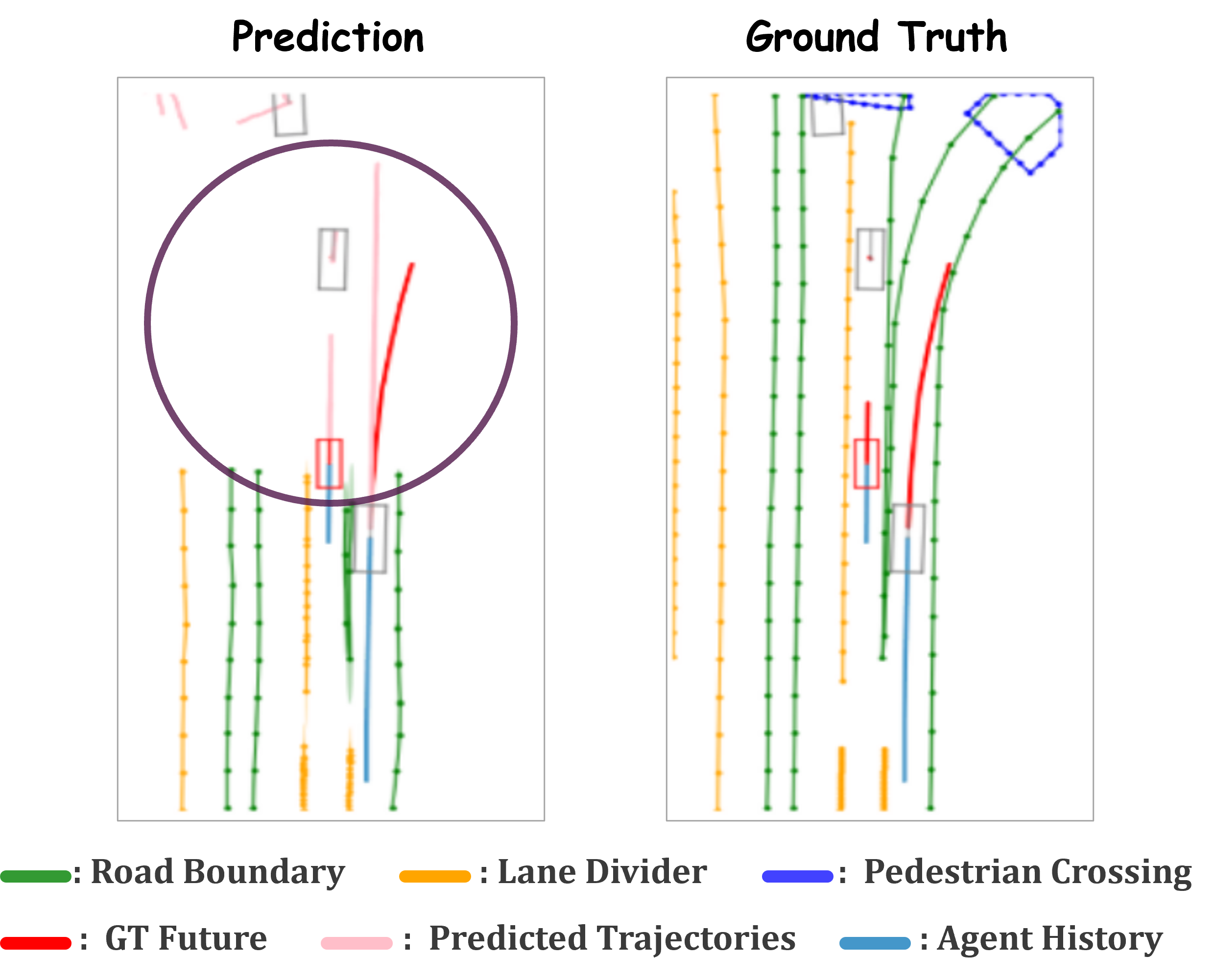}
\caption{The visual impact of inaccuracies in forward map elements on downstream tasks is demonstrated in the left subfigure. Due to the autonomous driving system's failure to correctly construct the ramp elements ahead of the vehicle, the system subsequently generated a dangerous straight-path trajectory.}
\label{supfig:downstream_task}
\vspace{-0.2in}
\end{figure}

\begin{figure}[t]
\centering
\setlength{\abovecaptionskip}{0.05cm}
\setlength{\belowcaptionskip}{-0.1cm}
\includegraphics[width=0.90\linewidth]{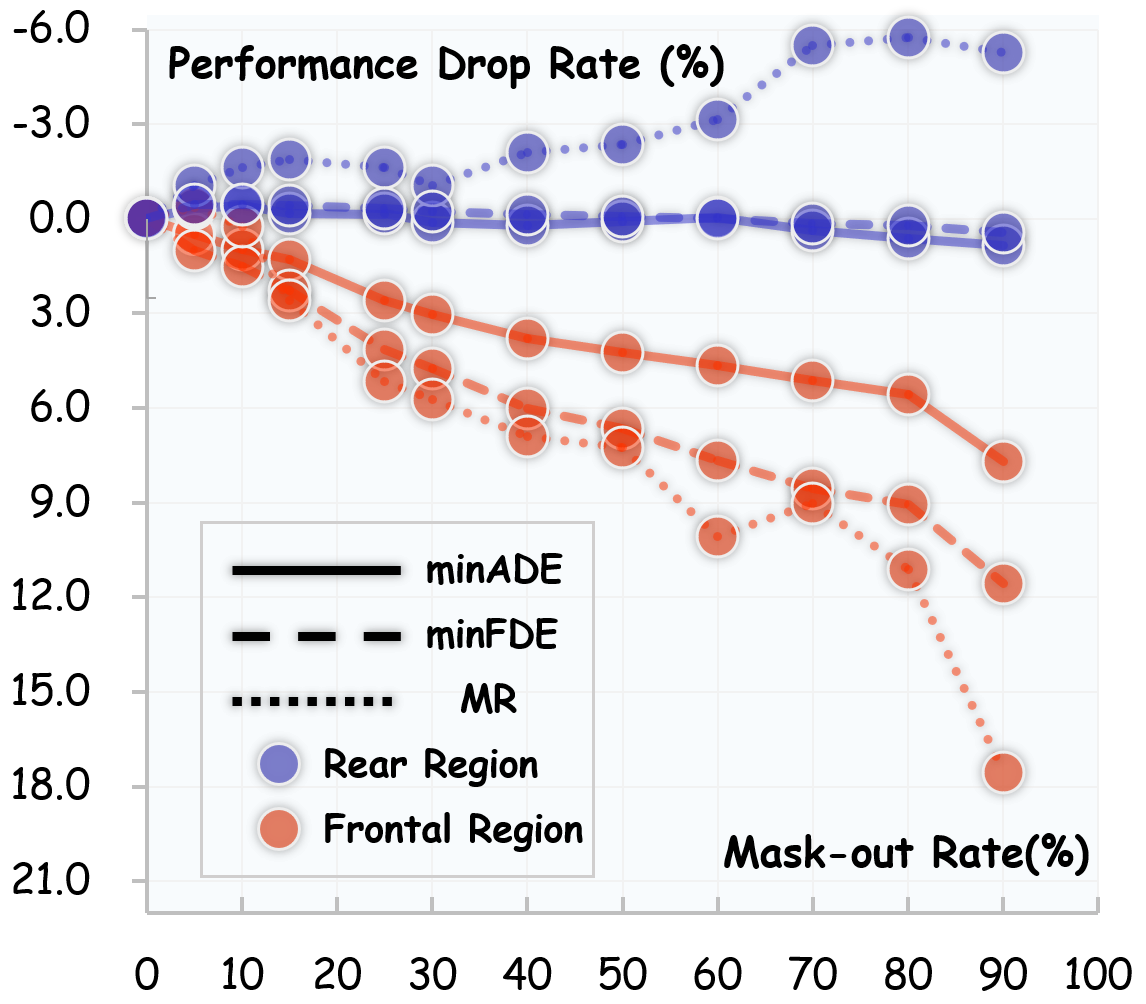}
\caption{\textbf{Impact of directional masking the outputs of the online HD mapper on downstream trajectory prediction.} As the Forward Mask ratio increases (red line), metrics such as minADE, minFDE, and MR deteriorate significantly, whereas the Backward Mask (blue line) has negligible impact. This empirical evidence demonstrates that downstream tasks are critically dependent on the ahead-aware mapping outputs. }
\label{supfig:downstream_bias}
\end{figure}

A key motivation behind the proposed AMap framework stems from our observation that map elements in front of the vehicle generally exert greater influence on downstream tasks, as illustrated in Figure~\ref{supfig:downstream_task}. Consequently, preliminary experiments were conducted to qualitatively and quantitatively validate the significantly higher importance of forward map areas compared to rear areas for downstream performance. In the baseline work~\cite{zhang2025iros}, the map elements around the vehicle predicted by the map builder \cite{liao2022maptr,liao2025maptrv2}, along with the historical trajectory of the vehicle, are used as inputs to the trajectory prediction model~\cite{zhou2022hivt}. Building upon this, we investigate the impact of forward online mapping on downstream trajectory prediction. We apply masks to different regions of the predicted map to simulate degradation in the prediction of map elements across various areas. As shown in Table~\ref{suptab:pre-downstream_forward}, Table~\ref{suptab:pre-downstream_backward}, and Figure~\ref{supfig:downstream_bias}, the experimental results reveal a clear and critical finding: degradation in forward map regions significantly impairs the performance of the downstream task, whereas degradation in rear map regions has almost no effect. Moreover, masking certain rear map elements can even lead to improved performance in the downstream task.

%\vspace{-2mm}
\begin{table}[htbp]
\centering
\caption{\textbf{Downstream Sensitivity to Forward Degradation.}}
\vspace{-2mm}
\fontsize{7pt}{9pt}\selectfont
\label{suptab:pre-downstream_forward}
\begin{tabular}[t]{c|ccc}
\toprule
\rowcolor{mygray}
 & \multicolumn{3}{c}{\bf Downstream Task Metric} \\
\cline{2-4}
\rowcolor{mygray}
\multirow{-2}{*}{\makecell[c]{\bf Mask \\ \bf Ratio}} & 
\multirow{-1}{*}{\bf minADE$\downarrow$} &
\multirow{-1}{*}{\bf minFDE$\downarrow$} &
\multirow{-1}{*}{\bf MR$\downarrow$} \\
\midrule
 - & 0.3856 & 0.7919 & 0.0854 \\
\midrule

5\% & 0.3875 \newgain{0.49\%} & 0.7998 \newgain{1.00\%} & 0.0851 \newdrop{0.35\%} \\
10\% & 0.3893 \newgain{0.96\%} & 0.8040 \newgain{1.53\%} & 0.0856 \newgain{0.23\%} \\
15\% & 0.3906 \newgain{1.30\%} & 0.8099 \newgain{2.27\%} & 0.0876 \newgain{2.58\%} \\
25\% & 0.3956 \newgain{2.59\%} & 0.8246 \newgain{4.12\%} & 0.0898 \newgain{5.15\%} \\
30\% & 0.3973 \newgain{3.03\%} & 0.8296 \newgain{4.76\%} & 0.0903 \newgain{5.74\%} \\
40\% & 0.4002 \newgain{3.79\%} & 0.8395 \newgain{6.01\%} & 0.0913 \newgain{6.91\%} \\
50\% & 0.4019 \newgain{4.23\%} & 0.8446 \newgain{6.65\%} & 0.0916 \newgain{7.26\%} \\
60\% & 0.4035 \newgain{4.64\%} & 0.8525 \newgain{7.65\%} & 0.0940 \newgain{10.07\%} \\
70\% & 0.4054 \newgain{5.13\%} & 0.8595 \newgain{8.54\%} & 0.0931 \newgain{9.02\%} \\
80\% & 0.4071 \newgain{5.58\%} & 0.8637 \newgain{9.07\%} & 0.0949 \newgain{11.12\%} \\
90\% & 0.4153 \newgain{7.70\%} & 0.8835 \newgain{11.57\%} & 0.1004 \newgain{17.56\%} \\
100\% & 0.4231 \newgain{9.73\%} & 0.9024 \newgain{13.95\%} & 0.1033 \newgain{20.96\%}  \\
\bottomrule
\end{tabular}
\vspace{-0.1in}
\end{table}

%\vspace{-2mm}
\begin{table}[htbp]
\centering
\caption{\textbf{Downstream Sensitivity to Backward Degradation.}}
\vspace{-2mm}
\fontsize{7pt}{9pt}\selectfont
\label{suptab:pre-downstream_backward}
\begin{tabular}[t]{c|ccc}
\toprule
\rowcolor{mygray}
 & \multicolumn{3}{c}{\bf Downstream Task Metric} \\
\cline{2-4}
\rowcolor{mygray}
\multirow{-2}{*}{\makecell[c]{\bf Mask \\ \bf Ratio}} & 
\multirow{-1}{*}{\bf minADE$\downarrow$} &
\multirow{-1}{*}{\bf minFDE$\downarrow$} &
\multirow{-1}{*}{\bf MR$\downarrow$} \\
\midrule
 - & 0.3856 & 0.7919 & 0.0854 \\
\midrule

5\% & 0.3843 \newdrop{0.34\%} & 0.7884 \newdrop{0.44\%} & 0.0845 \newdrop{1.05\%} \\
10\% & 0.3840 \newdrop{0.41\%} & 0.7884 \newdrop{0.44\%} & 0.0840 \newdrop{1.64\%} \\
15\% & 0.3849 \newdrop{0.18\%} & 0.7886 \newdrop{0.42\%} & 0.0838 \newdrop{1.87\%} \\
25\% & 0.3850 \newdrop{0.16\%} & 0.7892 \newdrop{0.34\%} & 0.0840 \newdrop{1.64\%} \\
30\% & 0.3860 \newgain{0.10\%}  & 0.7901 \newdrop{0.23\%} & 0.0845 \newdrop{1.05\%} \\
40\% & 0.3864 \newgain{0.21\%} & 0.7909 \newdrop{0.13\%} & 0.0836 \newdrop{2.11\%} \\
50\% & 0.3859 \newgain{0.08\%}  & 0.7913 \newdrop{0.08\%} & 0.0834 \newdrop{2.34\%} \\
60\% & 0.3855 \newdrop{0.03\%} & 0.7919 \newdrop{0.00\%} & 0.0827 \newdrop{3.16\%} \\
70\% & 0.3870 \newgain{0.36\%} & 0.7932 \newgain{0.16\%} & 0.0807 \newdrop{5.50\%} \\
80\% & 0.3880 \newgain{0.62\%} & 0.7936 \newgain{0.21\%} & 0.0805 \newdrop{5.74\%} \\
90\% & 0.3889 \newgain{0.86\%} & 0.7953 \newgain{0.43\%} & 0.0809 \newdrop{5.27\%} \\
100\% & 0.3882 \newgain{0.67\%} & 0.7971 \newgain{0.66\%} & 0.0816 \newdrop{4.45\%} \\

\bottomrule
\end{tabular}
\vspace{-0.1in}
\end{table}

\section{Comprehensive Experiments and Qualitative Results}
\label{sup_sec:2_ablation_and_vis}

%\TODO{bev mask ablation + nus vis(one page) + av2 vis(one page)}

\begin{table}[htbp]
\centering
\caption{\textbf{Impact of BEV Level Distillation.}}
\fontsize{7pt}{9pt}\selectfont
\label{suptab:ablation_bev}
\begin{tabular}[t]{l|l|cccc}
\toprule
\rowcolor{mygray}
\bf Group ID & 
\bf Method & 
\bf AP$_{p.}\uparrow$ &
\bf AP$_{d.}\uparrow$ &
\bf AP$_{b.}\uparrow$ &
\bf mAP$\uparrow$ 
\\
\midrule
\multirow{3}{*}{\bf Group \#~1} 
& baseline \#~1 & 71.67 & 67.26 & 69.12 & 69.35 \\
 &  + BEV KD & 68.96 & 71.45 & 68.89 & 69.76 \\
 &  + BEV Mask & 70.76 & 70.07 & 69.25 & 70.03 \\

\midrule

\multirow{3}{*}{\bf Group \#~2} 
& baseline \#~2 & 61.01 & 65.28 & 62.14 & 62.81 \\
 &  + BEV KD & 61.53 & 65.39 & 61.34 & 62.76 \\
 &  + BEV Mask & 62.49 & 65.18 & 62.09 & 63.25 \\

% \midrule

% \multirow{3}{*}{\bf Group \#~3} 
% & baseline \#~3 & x & x & x & 62.81 \\
%  &  + BEV KD & 61.62 & 64.72 & 61.62 & 62.66 \\
%  &  + BEV Mask & 62.67 & 66.10 & 61.64 & 63.47 \\

\bottomrule
\end{tabular}
\end{table}

\noindent\textbf{Ablation Study on BEV Level Distillation.} Two sets of ablation studies on BEV-level distillation are presented in Table~\ref{suptab:ablation_bev}. In Group \#~1, both teacher and student models are built upon the MapTracker~\cite{chen2024maptracker} codebase with 5-frame historical temporal fusion. The key difference lies in the backbone: the teacher employs ResNet-50~\cite{he2016resnet}, while the student uses ResNet-18~\cite{he2016resnet}. After applying basic BEV distillation, experimental results indicate performance degradation in two map element types (besides dividers) to varying degrees. We attribute this to differences in how map elements are represented in BEV feature space. Visualization of BEV features reveals that divider elements typically exhibit two characteristics: short length and longitudinal distribution, whereas crosswalk elements mostly appear as circular or transverse patterns, and boundary elements tend to be elongated. We hypothesize that the BEV representations of crosswalks and boundaries may introduce additional background noise during distillation, leading to performance degradation. To address this, we introduce a BEV mask that leverages ground truth map shapes to guide the student model to focus more on foreground representations while minimizing learning of background noise.

In Group \#~2, the teacher model remains the ResNet-50-based MapTracker with 5-frame temporal fusion, while the student model uses only the current frame with a ResNet-18 backbone. Compared to Group \#~1, the crosswalk element performance in this group differs from the baseline. BEV feature visualizations suggest that temporally fused BEV representations contain slight noise along the lateral dimension, possibly due to the dependence of temporal fusion on vehicle pose information, which may introduce lateral errors and consequently affect BEV feature accuracy in that direction. Both experimental groups demonstrate that BEV distillation with masking can maximize student model performance by effectively preserving critical structural information while suppressing noise.

\begin{figure*}[t!]
\begin{center}
   \includegraphics[width=0.94\linewidth]{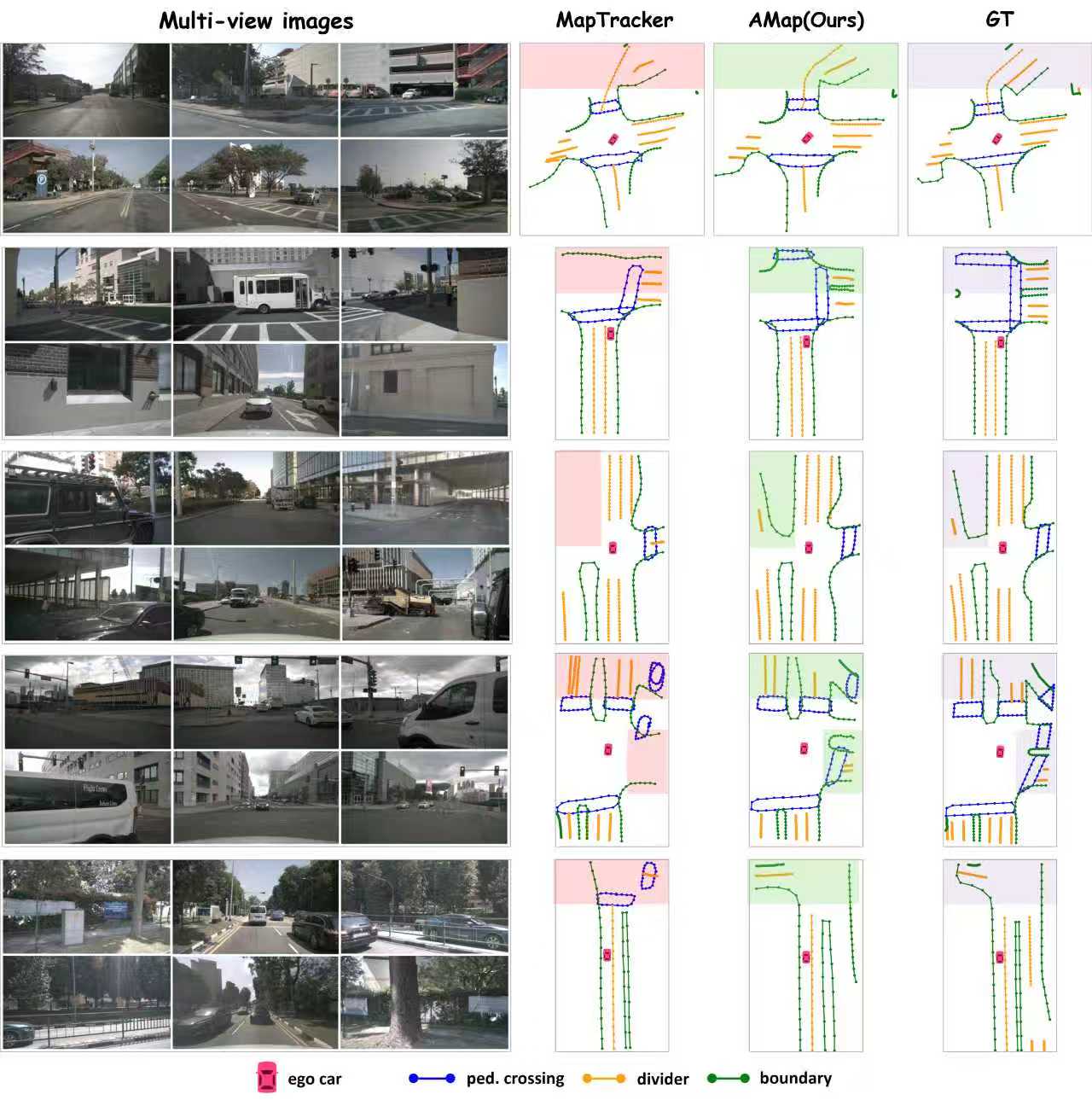}
\end{center}
   \caption{
   \textbf{Qualitative results on nuScenes dataset (Part 1).} From left to right: input multi-view images, MapTracker predictions, our predictions, and GT annotation. Each row corresponds to one sample. Shades of \colorbox{visred}{red} and \colorbox{visgreen}{green} are used to identify the differing and matching parts compared to the ground truth, which is represented by shade of \colorbox{visviolet}{violet}.
   }
\label{fig:sup_vis_nus_1}
\end{figure*}

\begin{figure*}[t!]
\begin{center}
   \includegraphics[width=0.94\linewidth]{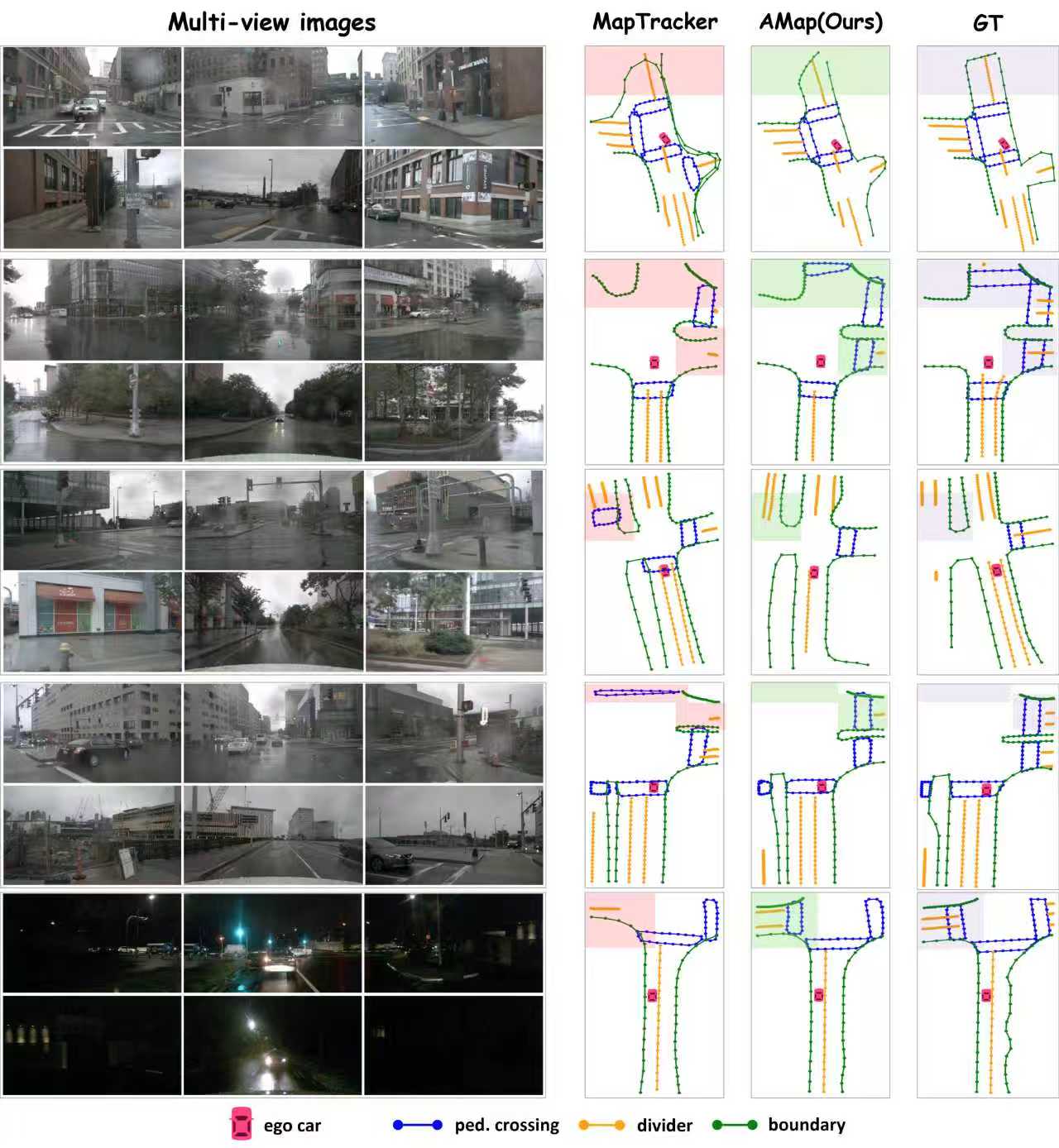}
\end{center}
   \caption{
   \textbf{Qualitative results on nuScenes dataset (Part 2).} From left to right: input multi-view images, MapTracker predictions, our predictions, and GT annotation. Each row corresponds to one sample. Shades of \colorbox{visred}{red} and \colorbox{visgreen}{green} are used to identify the differing and matching parts compared to the ground truth, which is represented by shade of \colorbox{visviolet}{violet}. Our proposed AMap demonstrates outstanding performance under various weather and lighting conditions, particularly in capturing the complex topological structure of map elements.
   }
\label{fig:sup_vis_nus_2}
\end{figure*}

\begin{figure*}[t!]
\begin{center}
   \includegraphics[width=0.78\linewidth]{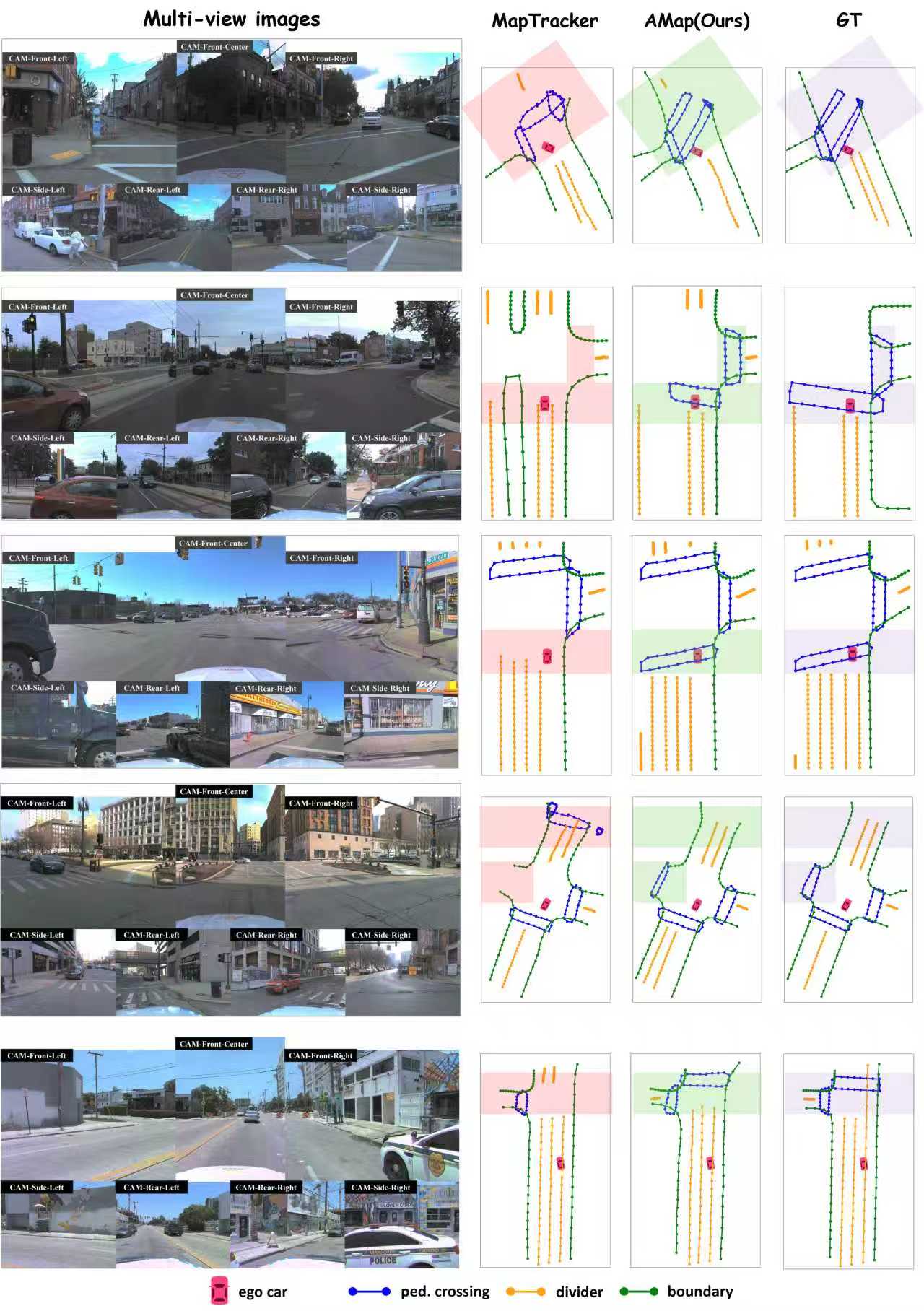}
\end{center}
   \caption{
   \textbf{Qualitative results on Argoverse 2 dataset.} From left to right: input multi-view images, MapTracker predictions, our predictions, and GT annotation. Each row corresponds to one sample. Shades of \colorbox{visred}{red} and \colorbox{visgreen}{green} are used to identify the differing and matching parts compared to the ground truth, which is represented by shade of \colorbox{visviolet}{violet}.
   }
\label{fig:sup_vis_av2}
\end{figure*}

\noindent\textbf{Additional Qualitative Experiments Results}
To visually demonstrate the effectiveness of our online map construction method, we conduct a qualitative visualization analysis on two autonomous driving benchmark datasets: nuScenes~\cite{caesar2020nuscenes} and Argoverse 2~\cite{wilson2023argoverse}.
As Shown in Figure~\ref{fig:sup_vis_nus_1}, Figure~\ref{fig:sup_vis_nus_2}, and Figure~\ref{fig:sup_vis_av2}, after applying our method, the quality of online map reconstruction in the forward region is significantly improved.

% \begin{figure*}[t!]
% \begin{center}
%    \includegraphics[width=0.80\linewidth]{sup_figures/av2_vis_2.jpg}
% \end{center}
%    \caption{
%    \textbf{Qualitative results on Argoverse 2.} From left to right: input multi-view images, MapTracker predictions, our predictions, and GT annotation. Each row corresponds to one sample. Shades of \colorbox{visred}{red} and \colorbox{visgreen}{green} are used to identify the differing and matching parts compared to the ground truth, which is represented by shade of \colorbox{visviolet}{violet}.
%    }
% \label{fig:sup_vis_av2_2}
% \end{figure*}

\section{Details of Experiments}
\label{sup_sec:3_details}

\noindent\textbf{Training Loss.} The overall loss function comprises two components: the basic loss $\mathcal{L}_{basic}$ computed against ground truth annotations, and a distillation loss derived from interaction with the teacher model. The basic loss~$\mathcal{L}_{\text{basic}}$ follows the same configuration as standard MapTracker~\cite{chen2024maptracker}, including BEV loss~$\mathcal{L}_{\text{BEV}}$, VEC loss~$\mathcal{L}_{\text{VEC}}$, and Transformation loss~$\mathcal{L}_{\text{trans}}$, which can be formally expressed as:
\begin{equation}
    \mathcal{L}_{\text{basic}} = \mathcal{L}_{\text{BEV}} + \mathcal{L}_{\text{VEC}} + \mathcal{L}_{\text{trans}}.
\end{equation}
For specific implementation details, please refer to the official settings of MapTracker~\cite{chen2024maptracker}.

The distillation loss consists of two parts: BEV feature loss~$\mathcal{L}_{\text{feat}}$ and Asymmetric Query loss~$\mathcal{L}_{\text{logitsKD}}$, formulated as:
\begin{equation}
    \mathcal{L}_{\text{distill}} = \mathcal{L}_{\text{feat}} + \mathcal{L}_{\text{logitsKD}}.
\end{equation}

The overall loss function can be formulated as:
\begin{equation}
    \mathcal{L}_{\text{all}} = \mathcal{L}_{\text{basic}}  
    + \mathcal{L}_{\text{distill}}.
\end{equation}

\noindent\textbf{Training Setting.} All experiments were conducted on 8 NVIDIA H20 GPUs. Unlike MapTracker's three-stage training paradigm (BEV pretrain, warm-up, and joint fine-tuning), the student model in AMap only undergoes the final training stage. We used a batch size of 16 and the AdamW optimizer with an initial learning rate of 5e-4. The training adopted a cosine learning rate scheduler with a final learning rate of 1.5e-6, and the weight decay was set to 0.01. In this work, we employ a consistent weighting scheme across all our experiments. Specifically, the weight for every loss term utilized throughout this paper is uniformly set to 1.0.

\begin{table}[htbp]
\centering
\caption{\textbf{Computational efficiency.}}
\fontsize{7pt}{9pt}\selectfont
\label{suptab:Computational_cost}
\begin{tabular}[t]{l|c|ccc}
\toprule
\rowcolor{mygray}
\bf Method & 
\bf Setting & 
\bf Params$\downarrow$ &
\bf Memory$\downarrow$ &
\bf FPS$\uparrow$ 
\\
\midrule
baseline & Train & \textbf{60.69~M} & \textbf{13.19~GB} & - \\
baseline & Test & \textbf{60.69~M} & \textbf{13.19~GB} & \textbf{31.5} \\

\midrule
baseline + Ours & Train & 118.4~M & 14.28~GB & - \\
baseline + Ours & Test & \textbf{60.69~M} & \textbf{13.19~GB} & \textbf{31.5} \\

\bottomrule
\end{tabular}
\end{table}

\begin{table*}[htbp]
\setlength{\tabcolsep}{3.0pt}
\setlength{\abovecaptionskip}{0.1cm}
\centering
\caption{\textbf{Benchmark Results on the nuScenes validation set.}}
\fontsize{7pt}{9pt}\selectfont
\label{suptab:benchmark}
\begin{tabular}[t]{lcccc|cccc|cccc|c|c}
\toprule
\rowcolor{mygray}
\bf Method & 
\bf Temporal &
\bf Backbone &
\bf BEV Encoder &
\bf Epoch &
\bf AP$_{p.}\uparrow$ &
\bf AP$_{d.}\uparrow$ &
\bf AP$_{b.}\uparrow$ &
\bf mAP$\uparrow$ &
\bf A-AP$_{p.}\uparrow$ &
\bf A-AP$_{d.}\uparrow$ &
\bf A-AP$_{b.}\uparrow$ &
\bf A-mAP$\uparrow$ &
\bf R-mAP$\uparrow$ 
\\
\midrule
\multirow{8}{*}{MapTR} & \ding{55} & R18 & GKT & 24 & 26.06 & 33.68 & 37.32 & 32.35 & 29.84 & 37.06 & 42.11 & 36.34 &  32.76 \\
& \ding{55} & R18 & GKT & 110 & 40.85 & 50.04 & 47.43 & 46.11 & 41.91 & 53.60 & 51.18 & 48.90 &  47.05 \\
& \ding{55} & R50 & GKT & 24 & 36.18 & 48.44 & 47.67 & 44.10 & 35.82 & 49.48 & 49.73 & 45.01 & 48.87 \\
& \ding{55} & R50 & GKT & 110 & 44.42 & 54.63 & 52.52 & 50.53 & 46.20 & 62.86 & 62.29 & 57.11 & 51.19 \\
& \ding{55} & R50 & BEVFormer & 24 & 34.53 & 45.43 & 44.88 & 41.61 & 35.67 & 47.86 & 49.14 & 44.22 & 45.89 \\
& \ding{55} & R50 & BEVPool & 24 & 44.96 & 51.86 & 53.48 & 50.10 & 46.80 & 56.69 & 58.61 & 54.03 &  46.30\\

& \ding{51} & R50 & GKT & 24 & 45.99 & 53.39 & 54.46 & 51.28 & 43.64 & 54.89 & 58.14 & 52.23 & 55.22 \\
& \ding{51} & R50 & BEVFormer & 24 & 51.71 & 54.88 & 53.29 & 53.29 & 48.80 & 56.86 & 55.41 & 53.69 & 55.18  \\

\midrule
\multirow{2}{*}{MapTRv2} & \ding{55} & R18 & BEVPool & 24 & 54.07 & 59.12 & 58.25 & 57.15 & 56.69 & 62.10 & 63.65 & 60.81 & 56.17 \\
& \ding{55} & R50 & BEVPool & 24 & 60.18 & 61.33 & 62.62 & 61.38 & 60.42 & 64.78 & 67.44 & 64.21 & 61.82\\

\midrule
\multirow{3}{*}{MapQR} & \ding{55} & R18 & BEVFormer & 24 & 59.33 & 63.26 & 64.35 & 62.31 & 60.03 & 66.09 & 67.98 & 64.70 & 63.13\\
& \ding{55} & R50 & BEVFormer & 24 & 63.36 & 68.03 & 67.67 & 66.35 & 62.47 & 71.20 & 71.47 & 68.38 & 65.59\\
& \ding{55} & R50 & BEVFormer & 110 & 65.42 & 76.42 & 77.53 & 73.12 & 69.12 & 77.57 & 76.45 & 74.38 & 72.57 \\

\midrule
\multirow{1}{*}{MGMap} & \ding{55} & R50 & BEVFormer & 24 & 48.21 & 55.80 & 55.43 & 53.15 & 48.56 & 57.21 & 59.64 & 55.13 & 56.78 \\

\midrule
\multirow{5}{*}{GeMap} & \ding{55} & R50 & BEVFormer & 24 & 45.62 & 53.95 & 54.20 & 51.26 & 41.37 & 56.91 & 56.56 & 51.61 & 54.74\\
& \ding{55} & R50 & BEVFormer & 110 & 59.80 & 65.08 & 63.19 & 62.69 & 58.26 & 68.47 & 68.44 & 65.06 & 63.91  \\
& \ding{55} & Swin-T & BEVFormer & 110 & 70.43 & 72.77 & 72.76 & 71.99 & 69.89 & 75.24 & 76.18 & 73.77 & 72.79 \\
& \ding{55} & V2-99 & BEVFormer & 110 & 70.18 & 72.19 & 73.70 & 72.02 & 68.97 & 75.16 & 76.44 & 73.52 & 72.66 \\
& \ding{55} & V-99* & BEVFormer & 110 & 74.28 & 75.99 & 77.68 & 75.99 & 72.70 & 78.00 & 80.24 & 76.98 & 76.36 \\

\midrule
\multirow{2}{*}{HRMapNet} & \ding{55} & R50 & BEVFormer & 24 & 62.46 & 64.81 & 66.33 & 64.54 & 57.63 & 63.79 & 65.61 & 62.34 & 69.51 \\
& \ding{55} & R50 & BEVFormer & 110 & 70.52 & 70.77 & 74.27 & 71.85 & 62.98 & 70.53 & 73.89 & 69.13 & 77.28 \\

\bottomrule
\end{tabular}

\begin{tablenotes}
\item[1] \scriptsize{\textbullet\; AP$_{p.}$, AP$_{d.}$, AP$_{b.}$ denote the Average Precision (AP) for pedestrian crossings, lane dividers, and road boundaries, respectively.}
\end{tablenotes}

\vspace{-0.1in}
\end{table*}

\noindent\textbf{Computational Efficiency.} The computational overhead during training and inference is summarized in Table~\ref{suptab:Computational_cost}. While AMap incorporates additional parameters by loading the teacher model during training, this overhead can be mitigated by precomputing and caching intermediate tensors from the teacher model offline. During inference, the AMap framework achieves improved accuracy without introducing any additional computational cost.

\section{Enhanced Benchmark Evaluation}

\label{sup_sec:4_benchmark}

\begin{figure*}[tbh]
\centering
\setlength{\abovecaptionskip}{0.05cm}
\setlength{\belowcaptionskip}{-0.1cm}
\includegraphics[width=0.90\linewidth]{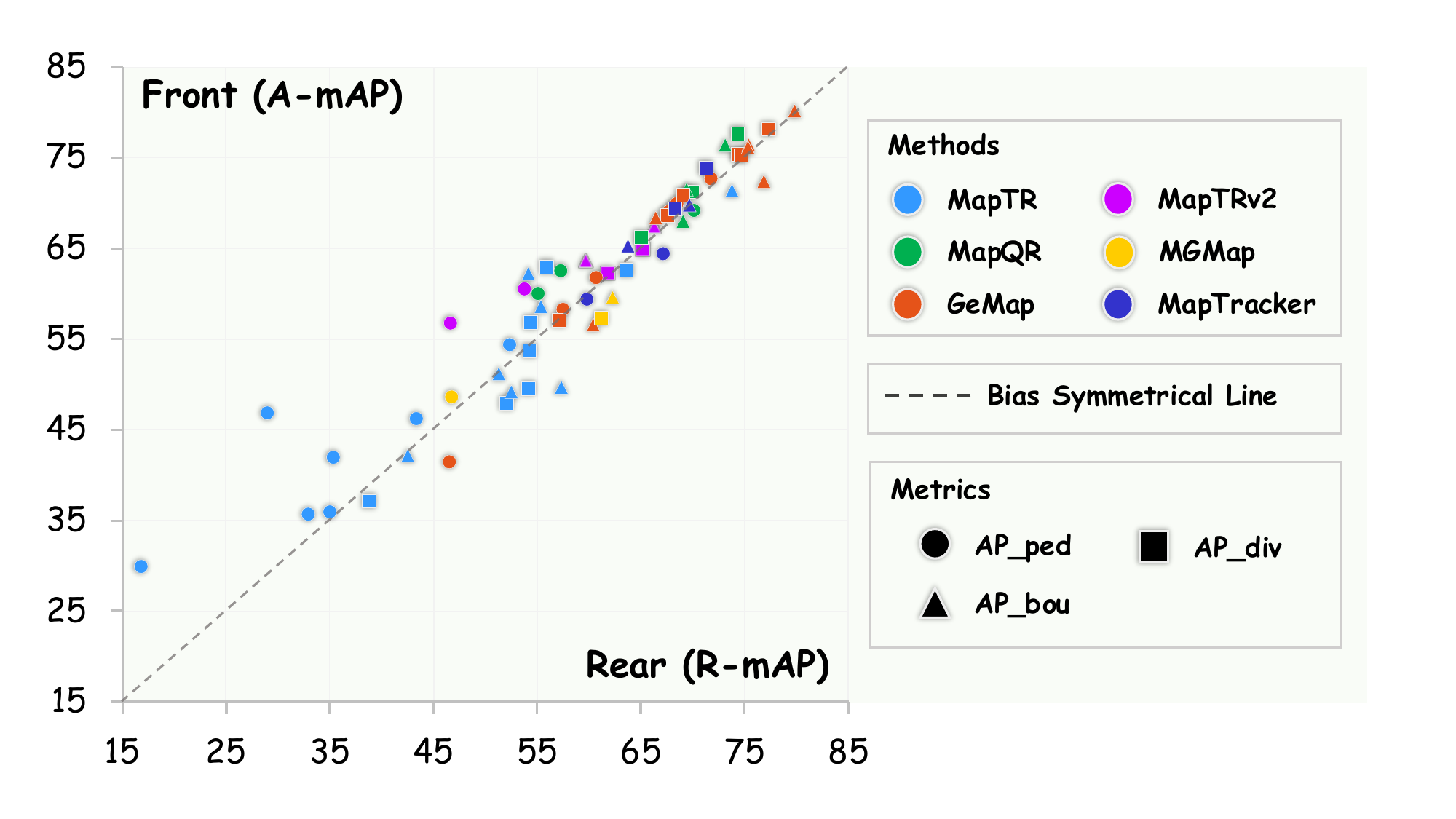}
\caption{\textbf{Front-Rear mAP Performance Benchmark for Single-Frame Models.} Compared to temporal models, which rely on history aggregation, single-frame models exhibit a significantly reduced tendency toward rear performance bias. This is evident as the data points primarily cluster tightly around the y=x reference line, indicating similar performance regardless of the direction.}
\label{supfig:bench}
\vspace{-0.2in}
\end{figure*}

% \TODO{benchmark + analysis}
We also benchmark the performance of various common single-frame models for online HDMap construction as shown in Table~\ref{suptab:benchmark}, including MapQR~\citep{liu2024mapqr}, GeMap~\cite{zhang2024gemap}, MapTRv2~\cite{liao2025maptrv2}, MGMap~\cite{liu2024mgmap}, and the single-frame variants of MapTR~\cite{liao2022maptr} and MapTracker~\cite{chen2024maptracker}, using the A-mAP and R-mAP metrics introduced in the main text. As illustrated in Fig~\ref{supfig:bench}, single-frame models do not suffer from the forward-perception deficit issue. Notably, a comparison between the single-frame and temporal variants (from main text) of MapTR and MapTracker clearly demonstrates that temporal aggregation introduces a significant bias towards superior performance in rear regions.

% \section{Rationale}
% \label{sec:rationale}
% % 
% Having the supplementary compiled together with the main paper means that:
% % 
% \begin{itemize}
% \item The supplementary can back-reference sections of the main paper, for example, we can refer to \cref{sec:intro};
% \item The main paper can forward reference sub-sections within the supplementary explicitly (e.g. referring to a particular experiment); 
% \item When submitted to arXiv, the supplementary will already included at the end of the paper.
% \end{itemize}
% % 
% To split the supplementary pages from the main paper, you can use \href{https://support.apple.com/en-ca/guide/preview/prvw11793/mac#:~:text=Delete%20a%20page%20from%20a,or%20choose%20Edit%20%3E%20Delete).}{Preview (on macOS)}, \href{https://www.adobe.com/acrobat/how-to/delete-pages-from-pdf.html#:~:text=Choose%20%E2%80%9CTools%E2%80%9D%20%3E%20%E2%80%9COrganize,or%20pages%20from%20the%20file.}{Adobe Acrobat} (on all OSs), as well as \href{https://superuser.com/questions/517986/is-it-possible-to-delete-some-pages-of-a-pdf-document}{command line tools}.

\clearpage

{
    \small
    \bibliographystyle{ieeenat_fullname}
    \bibliography{main}
}

\end{document}